\documentclass{article}
\usepackage{spconf,amsmath,epsfig}

\usepackage{mmstyles}
\usepackage{threeparttable}
\usepackage{array}
\usepackage{subfigure}
\usepackage{mmstyles}

\makeatletter
\def\hlinew#1{%
  \noalign{\ifnum0=`}\fi\hrule \@height #1 \futurelet
   \reserved@a\@xhline}
\makeatother

\begin{document}\sloppy

% Example definitions.
% --------------------
\def\x{{\mathbf x}}
\def\L{{\cal L}}

\renewcommand{\baselinestretch}{0.9}
% Title.
% ------
\title{A Large Scale Urban Surveillance Video Dataset for \\
Multiple-Object Tracking and Behavior Analysis}
%
% Single address.
% ---------------
% \name{Anonymous ICME submission (Paper ID 425)}

% \address{}
% % \address{Paper ID 425}
\name{ Guojun Yin, Bin Liu\thanks{Bin Liu is the corresponding author.}, Huihui Zhu, Tao Gong, Nenghai Yu}
\address{ \normalsize University of Science and Technology of China, \\
\normalsize Key Laboratory of  Electromagnetic Space Information, The Chinese Academy of Sciences, \\
\tt\small \{gjyin,zhuhui33,gt950513\}@mail.ustc.edu.cn, \{flowice,ynh\}@ustc.edu.cn
}

\maketitle

\begin{abstract}
Multiple-object tracking and behavior analysis have been the essential parts of surveillance video analysis for public security and urban management. 
With billions of surveillance video captured all over the world, multiple-object tracking and behavior analysis by manual labor are cumbersome and cost expensive.
Due to the rapid development of deep learning algorithms in recent years, automatic object tracking and behavior analysis put forward an urgent demand on a large scale well-annotated surveillance video dataset that can reflect the diverse, congested, and complicated scenarios in real applications. 
This paper introduces an urban surveillance video dataset (USVD) which is by far the largest and most comprehensive. 
The dataset consists of $16$ scenes captured in $7$ typical outdoor scenarios: street, crossroads, hospital entrance, school gate, park, pedestrian mall, and public square. 
Over $200k$ video frames are annotated carefully, resulting in more than $3.7$ million object bounding boxes and about $7.1$ thousand trajectories. 
We further use this dataset to evaluate the performance of typical algorithms for multiple-object tracking and anomaly behavior analysis and explore the robustness of these methods in urban congested scenarios.
\end{abstract}
%
% \begin{keywords}
% One, two, three, four, five
% \end{keywords}
%
%%%%%%%%% BODY TEXT
\vspace{-0.25cm}
\section{Introduction}
\label{section:introduction}
\vspace{-0.25cm}

With the rapid development of digital acquisition and storage technologies, video surveillance has become one of the most important safety monitoring methods widely used all around the world. 
As a very active research field in the computer vision area, the main goal of the research on the surveillance video is to effectively analyze and extract information from a large amount of unstructured video data acquired by the surveillance cameras, automatically detect, track and identify the targets, analyze various behaviors of the targets, understand various events occurring in the scene, and alarm suspicious events, to provide technical support for public security.

Among various research topics in surveillance video analysis and scene recognition, multiple-object tracking and behavior analysis is one of the major research fields.
After the proposition of the concepts of \textit{intelligent transportation} and \textit{smart city}, more and more researchers have begun to focus on object tracking and behavior analysis in the surveillance videos~\cite{CVPR09peds,wang2011automatic}.
However, the explosive growth of the number of vehicles and populations has resulted in more congested and complicated urban environments, which brings many new challenges to the research on the surveillance video.

As many object tracking and anomaly behavior analysis algorithms have been proposed to deal with the congested and complicated scenarios~\cite{bae2014robust,hong2016online,liu2016ssd,ren2015faster} , the corresponding public challenging datasets are required to provide the fair comparison. 
However, there are only several real-world urban surveillance video datasets serving the purpose of evaluating the performance and robustness of object tracking and behavior algorithms. 
And most of the existing surveillance video datasets [16] used in the previous works are relatively small and simple, which makes them less qualified to assess the performance in real-world applications for more and more congested and complex scenarios.

\begin{figure}[t]
\centering
\includegraphics[width=0.99\columnwidth]{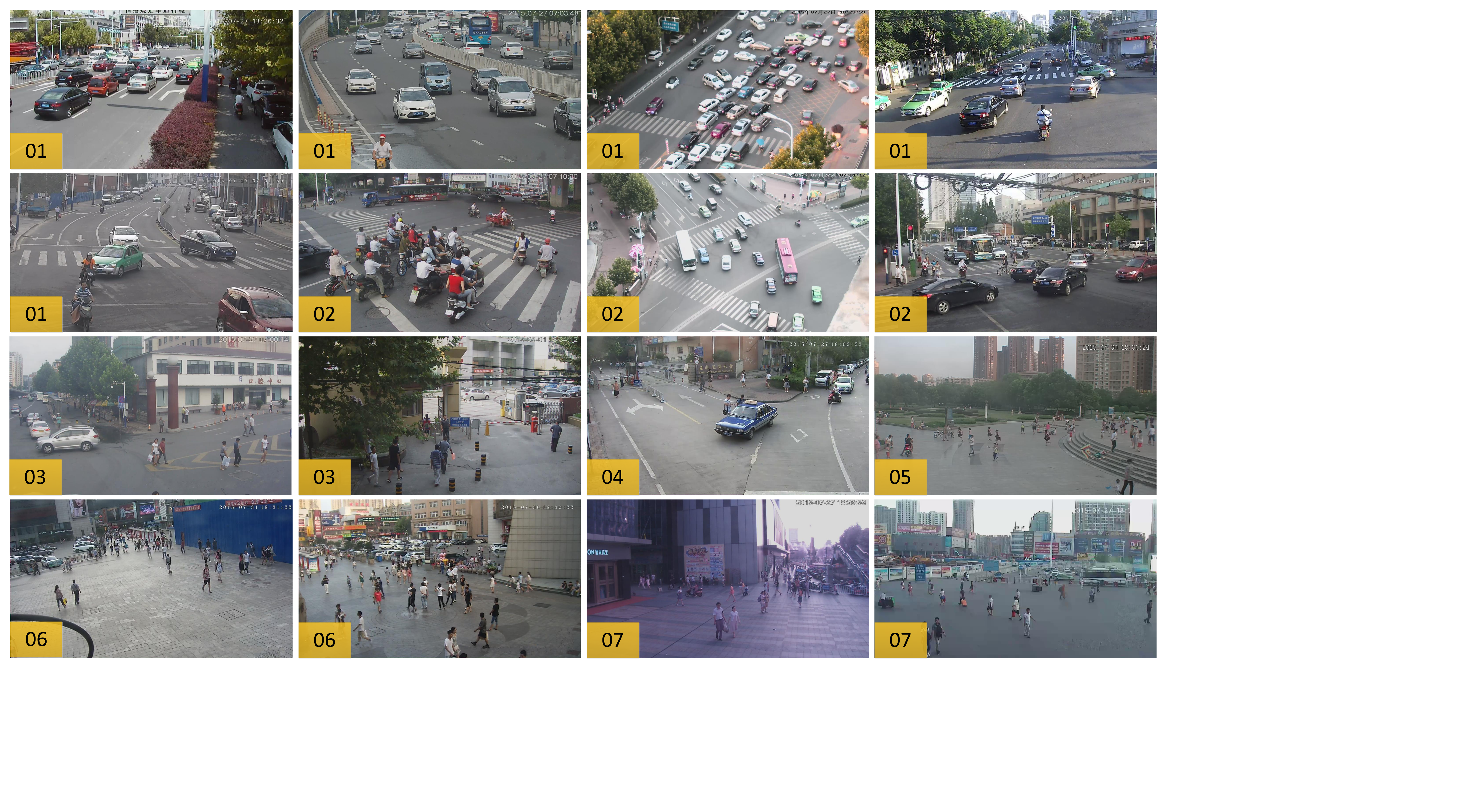}
\vspace{-0.35cm}
\caption{The congested traffic scenes in urban environments. The ID numbers in the corners represent the scenarios respectively: street, crossroads, hospital entrance, school gate, park, pedestrian mall and public square. }
\label{fig:scenes}
\vspace{-0.35cm}
\end{figure}

\begin{table*}[t]
\centering
\begin{threeparttable}
\footnotesize
\caption{Comparison of existing datasets in urban environments (object tracking and detection).}
\label{table:exist_dataset}
\begin{tabular}{l|r|r|r|r|r|r|l}
\hlinew{1.1pt}
Dataset & \#Clips & \#Annotated frames & \#Tracks & \#Boxes & \#Density & Camera & Task\\
\hline
MIT Traffic\cite{wang2011automatic} & 20 & 520 & - & 2054 & 3.95 & static & Pedestrian detection \\
\hline
Caltech Pedestrian\cite{CVPR09peds} & 11 & \textbf{250000} & - & 350000 & 1.40 & dynamic & Pedestrian detection \\
\hline
Daimler Pedestrian * \cite{ess2007depth} & 1 & 21790 & - & 56492 & 2.59 & dynamic & Pedestrian detection\\
\hline
KITTI Tracking ** \cite{Geiger2012CVPR} & 21 & 7924 & 917 & 8896 & 1.12 & dynamic & Multiple object tracking\\
\hline
MOT Challenge 2015 2D~\cite{leal2015motchallenge} & 22 & 11286 & 1221 & 101345 & 8.98 & diverse & Multiple object tracking \\
\hline
MOT Challenge 2016~\cite{milan2016mot16} & 14 & 11235 & 1342 & 292733 & 26.06 & diverse & Multiple object tracking \\
\hline
MOT Challenge 2017~\cite{milan2016mot16} *** & 14 & 11235 &  1331 & 300373 & \textbf{26.73} & diverse & Multiple object tracking \\
\hline
Our dataset & \textbf{32} & 211706 & \textbf{7173} & \textbf{3758821} & 17.75 & static & Object detection and tracking\\
\hlinew{1.1pt}
\end{tabular}

\begin{tablenotes}
\footnotesize
\item{*} The statistics of Daimler Pedestrian detection dataset only includes test split.
\item{**} The statistics of KITTI Tracking dataset only includes training split and boxes include the bounding boxes of \emph{DontCare} labels.
\item{***} The sequences in MOT17 Challenge are the same as MOT16 sequences with a new, more accurate ground truth.
\end{tablenotes}
\end{threeparttable}
\vspace{-0.5cm}
\end{table*}

In the current work, we propose a large scale urban surveillance video dataset (USVD)
with congested and complex scenarios for multiple-object tracking and anomaly behavior analysis. 
To the best of our knowledge, it is to-date the largest and most realistic
public dataset for real video surveillance. 
There are mainly four advantages in our dataset compared with the existing datasets.

\textbf{Realistic.} All the data are from the real public surveillance
    scenes, which enables the evaluation of computer vision algorithms
    on direct application to the real-world.

\textbf{Complex.} The dataset is comprised of $7$ typical
    scenarios with $16$ different congested scenes. There are frequent
    occlusion, deformation, various viewpoints and diverse targets in these congested scenes.

\textbf{Large Scale.} The dataset consists of over $200k$ annotated frames and
    more than $3.7$ million bounding boxes of about $7.1$ thousand unique trajectories.

\textbf{Well-Annotated.} All the bounding boxes are manually annotated
    and checked. The annotation includes location, size, object category,
    occlusion, and trajectory identity.

% \textbf{1) Realistic.} All the data are from the real public surveillance
%     scenes, which enables the evaluation of computer vision algorithms
%     on direct application to the real-world.
% %
% \textbf{2) Complex.} The dataset is comprised of $7$ typical
%     scenarios with $16$ different congested scenes. There are frequent
%     occlusion, deformation, various viewpoints and diverse targets in these congested scenes.
% %
% \textbf{3) Large Scale.} The dataset consists of over $200k$ annotated frames and
%     more than $3.7$ million bounding boxes of about $7.1$ thousand unique trajectories.
% %
% \textbf{4) Well-Annotated.} All the bounding boxes are manually annotated
%     and checked. The annotation includes location, size, object category,
%     occlusion, and trajectory identity.

We also use the proposed dataset to evaluate the performance
of typical algorithms  for multiple-object tracking and anomaly detection and explore the robustness of these methods in urban congested conditions.

% The rest of the paper is organized as follows. We briefly discuss the existing urban
% databases for tracking and detection in Section~\ref{section:related_work}. In Section~\ref{section:lsvsd}, we present the details of the large sclae urban surveillance video dataset including data collection, data annotation, data format and annotation statistics. In Sec.~\ref{section:application}, we introduce the applications of
% the proposed database and evaluate the performance of several object detection and multiple-object tracking algorithms. We conclude this paper in Section~\ref{section:conclusion}.

\vspace{-0.25cm}
\section{Related Works}
\label{section:related_work}
\vspace{-0.25cm}

In the recent years, the computer vision community has created benchmarks for video related tasks such as scene recognition, pedestrian \& object detection,
object tracking, action recognition, anomaly behavior detection and etc. 
Despite the potential pitfalls of such datasets, they have proved to be extremely helpful to advance the state-of-the-arts in the corresponding
areas~\cite{Everingham10,leal2015motchallenge,lin2014microsoft,ILSVRC15}.
An overview of examples of the existing datasets in urban environments for object detection and tracking is shown in Tab.~\ref{table:exist_dataset}.
% \if 0
% However, there are quite few
% datasets for tracking and detection in urban video surveillance scenarios
% and the related quantitative evaluation.
% \fi

\textbf{Real Urban Video Datasets.}
The MIT Traffic dataset~\cite{wang2011automatic}  is an example of the recent efforts to build more realistic urban traffic surveillance video datasets for research on pedestrian detection and activity analysis. 
It includes a traffic video sequence of 90 minutes long, recorded by a stationary camera and 
the whole sequence is divided into 20 clips. 
The size of the scene is 720 by 480. 
In order to evaluate the performance of human detection on this dataset, 
the ground truth of the pedestrians of some sampled frames are manually labeled. There are in total 520 annotated frames and 2054 bounding boxes in the dataset.

The Caltech Pedestrian dataset~\cite{CVPR09peds,Dollar2012PAMI} consists of approximately 10 hours of $640\times480$ $30$Hz video taken from a vehicle driving through the regular traffic in an urban environment.
All the data  is roughly divided in half, setting aside 6 sessions for training and 5 for testing. 
About $250000$ frames with a total of $350000$ bounding boxes and $2300$ unique pedestrians are annotated.

The Daimler Monocular Pedestrian Detection dataset~\cite{ess2007depth} is another  dataset for pedestrian detection in urban environments. The training set contains $15560$ pedestrian samples (image cut-outs at $48\times96$ resolution) and 6744 additional full images without pedestrians for extracting negative samples. 
The test set contains an independent sequence with more than $21790$ images and $56492$ pedestrian labels (fully visible or partially occluded), captured from a vehicle during a 27 min driving through the urban traffic.

Although they are helpful to evaluate
the performance of pedestrian detection algorithms in real surveillance video to
some extent, the three datasets mentioned above only include some uncrowded urban scenes and a
few annotated bounding boxes, which is relatively simple compared to the complex and congested traffic
conditions nowadays.

\textbf{MOT Datasets.} 
Recently, the KITTI benchmark~\cite{Geiger2012CVPR} is introduced for
challenges in autonomous driving, which includes stereo flow, odometry, road
and lane estimation, object detection as well as
tracking. Some of the sequences include crowded pedestrian crossings, making
the dataset quite challenging, but 
% the camera position is always the same for all sequences (at a car's height) and 
the camera is moving, while the conventional traffic surveillance video  varies greatly.

MOT challenge 2015~\cite{leal2015motchallenge}, challenge 2016~\cite{milan2016mot16} and challenge 2017~\cite{milan2016mot16} are the recent challenging benchmarks for multiple object tracking. The videos in the benchmarks are diverse, and some of which are selected from the existing datasets, \eg KITTI Tracking dataset. However, the dataset consists of  various video types including  surveillance videos and moving-camera videos. 
Therefore, it motivates us to establish a public challenging urban surveillance video dataset which
is more realistic to evaluate the performance of various algorithms for object tracking and behavior analysis.

\begin{figure*}[t]
\centering
\subfigure[Raw data]{
\includegraphics[width=0.49\columnwidth]{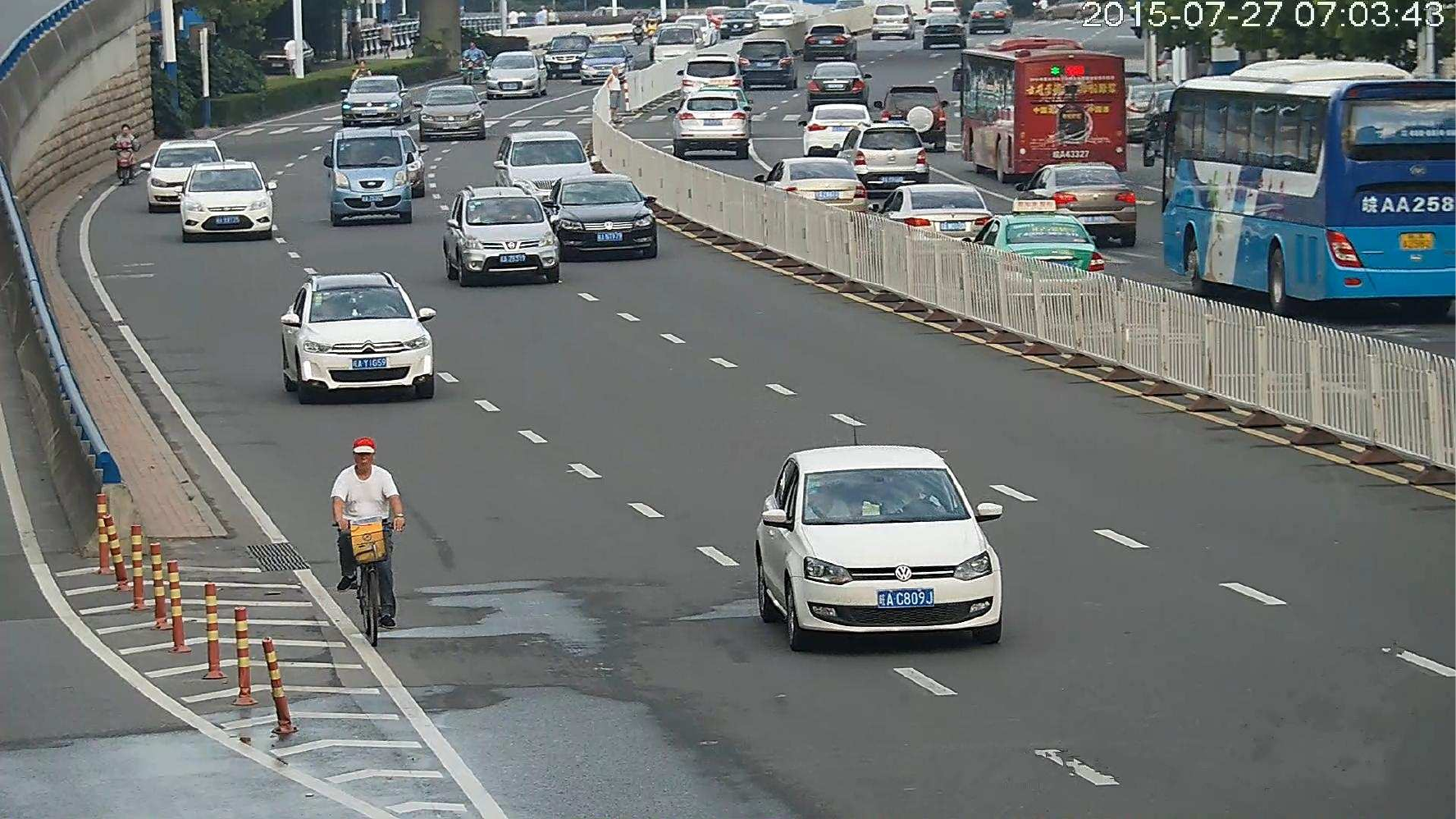}}
\subfigure[Ground truth]{
\includegraphics[width=0.49\columnwidth]{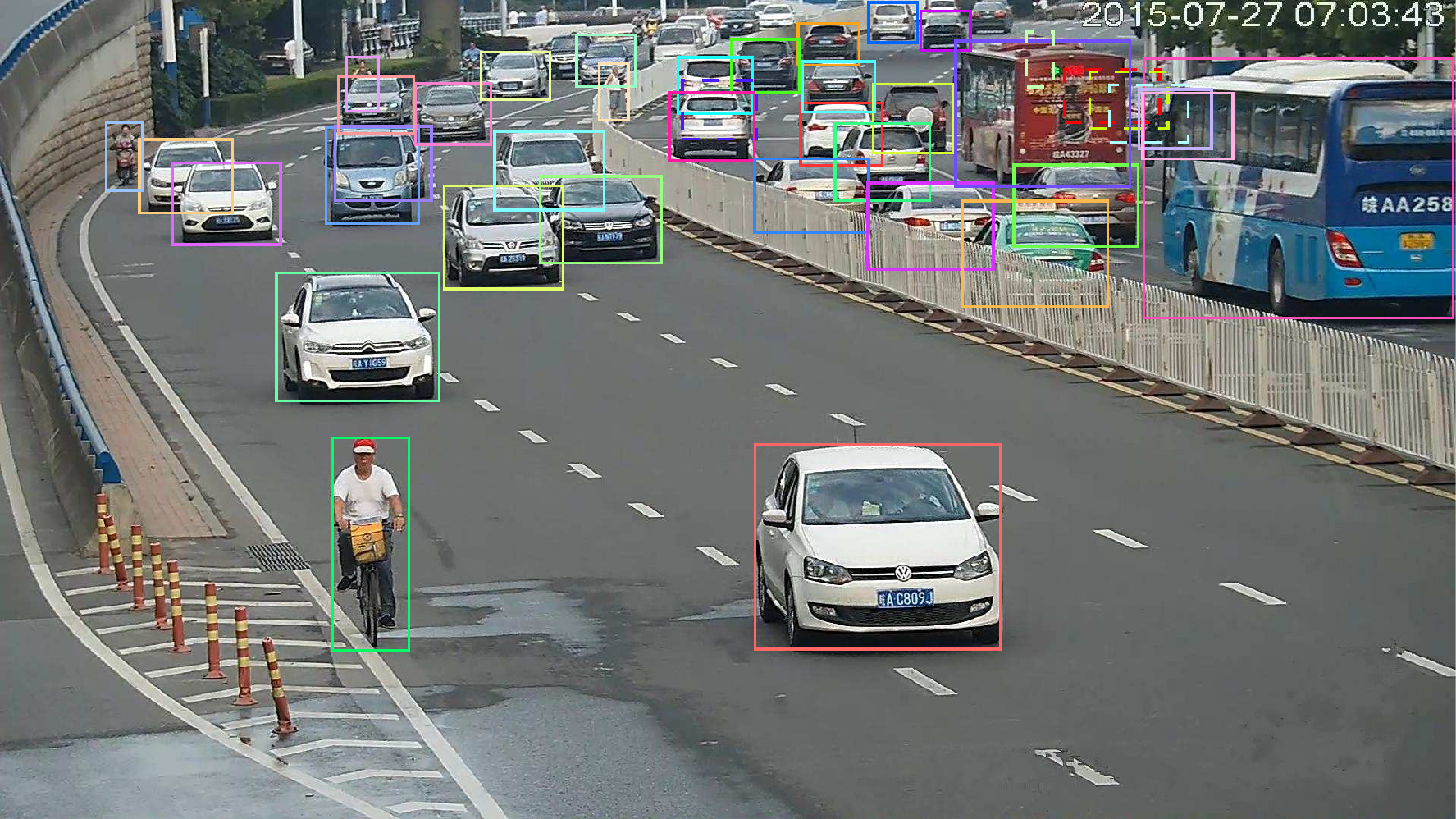}}
\subfigure[Detection results]{
\includegraphics[width=0.49\columnwidth]{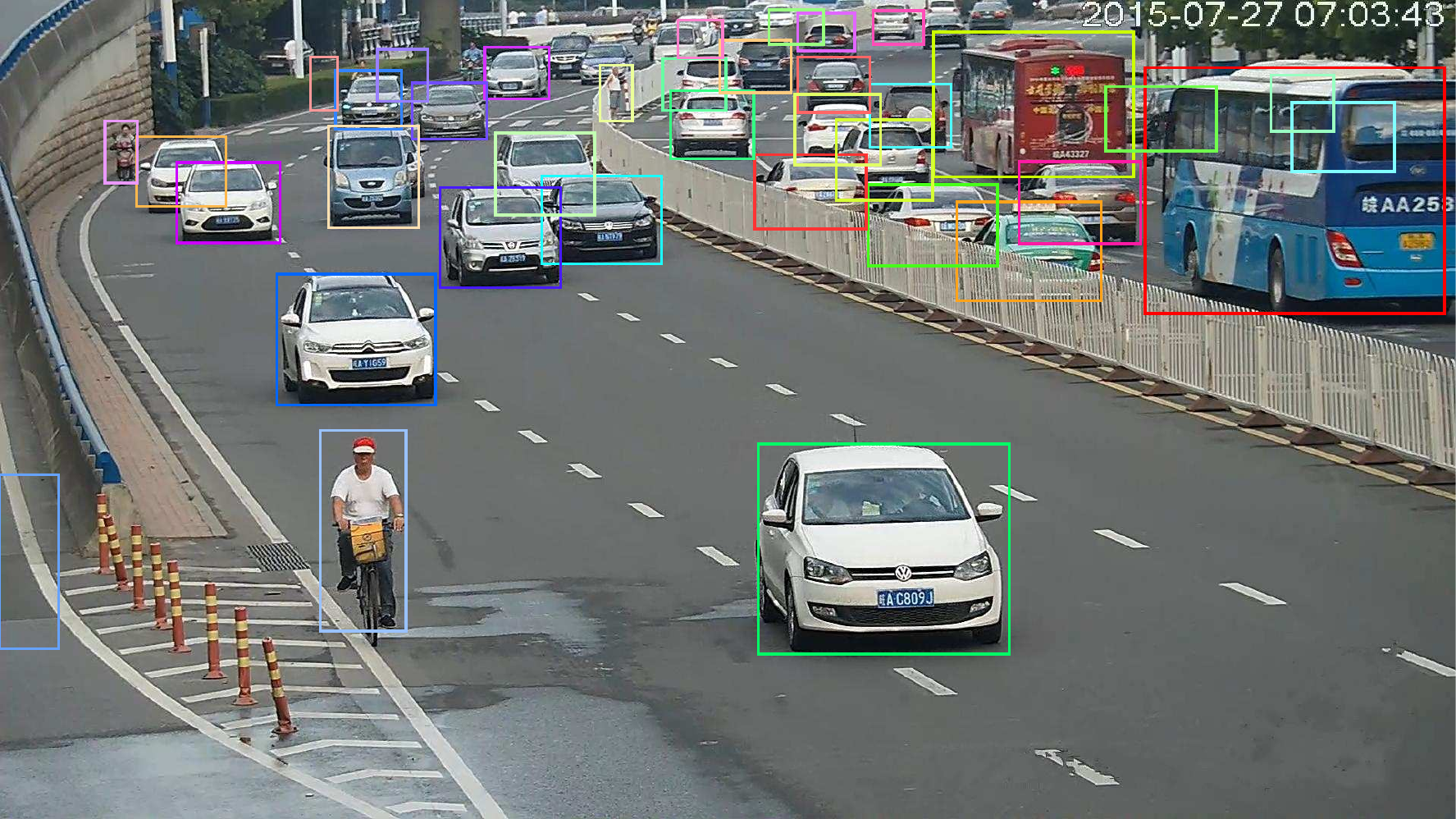}}
\subfigure[Trajectory distribution]{
\includegraphics[width=0.49\columnwidth]{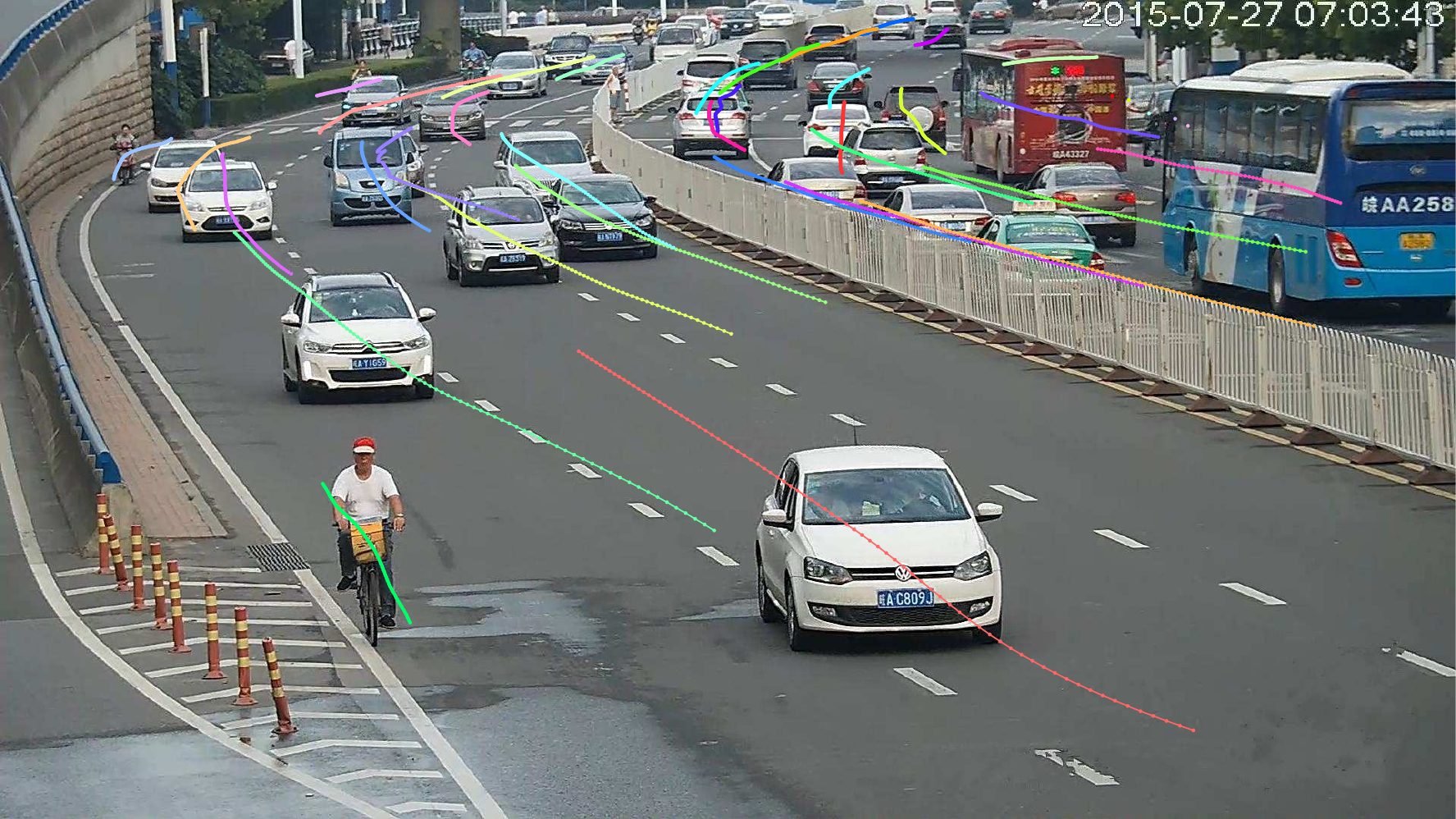}}
\vspace{-0.4cm}
\caption{An example sequence in the proposed USVD dataset.}
\label{fig:example_sequence}
\vspace{-0.4cm}
\end{figure*}

\vspace{-0.25cm}
\section{Large Scale Urban Surveillance Video Dataset}
\label{section:lsvsd}
\vspace{-0.25cm}

An example sequence in our proposed dataset is shown in Fig.~\ref{fig:example_sequence}, in which the trajectory distribution only includes the trajectory location in the adjacent $100$ frames~($4$ seconds) and the bounding boxes of dotted line mean the completely-occluded target.
The targets we are interested in urban environments are movable individuals or units, \eg, pedestrian, car or van, rather than the stationary objects, \eg, trees, pillars or traffic lights. 
% For example, we are interested in the traffic participants such as  but not interested in trees, pillars or traffic lights.
%
In this section, we introduce our large scale urban surveillance video dataset (USVD) in details. 
%
% We first introduce the data collection and then propose a clear protocol that is obeyed throughout the annotation of the entire dataset to guarantee the consistency. Finally we discuss and analyze the statistics of the proposed dataset.

\begin{table*}[!t]
\centering
\small
\caption{Statistics of sequences included in the proposed dataset. }
\label{table:sequences_included}

\begin{tabular}
{c|c|r|r|r|r|c|c|r|r|r|r}
\hlinew{1.1pt}
\multicolumn{6}{c|}{\textbf{Train sequences}} & \multicolumn{6}{c}{\textbf{Test sequences}}\\
\hline
Seq & Resolution & Length & Tracks & Boxes & Density & Seq & Resolution & Length & Tracks & Boxes & Density  \\
\hline
01 & $1280\times{720}$ & 10601 & 240 & 91055 & 8.5893 & 02 & $1280\times{720}$ & 10989 & 219 & 10989 & 10.7766 \\
03 & $1920\times{1080}$ & 11000 & 389 & 234806 & 21.3460 & 04 & $1920\times{1080}$ &11101& 460& 241924& 21.7930 \\
05 & $1920\times{1080}$ & 4500 & 500 & 246923 & 54.8718 & 06 & $1920\times{1080}$ & 3501 & 407 & 209580 & 59.8629 \\
07 & $1280\times{720}$ & 6001 & 178 & 67160 & 11.1915 & 08 & $1280\times{720}$ & 6001 & 174 & 71361 & 11.8915 \\
09 & $1280\times{720}$ & 6000 & 250 & 80973 & 13.4955 & 10 & $1280\times{720}$ & 7501 & 253 & 100381 & 13.3823 \\
11 & $1280\times{720}$ & 7500 & 363& 80416&10.7221 & 12 & $1280\times{720}$ & 7501 & 465& 92760 & 12.3664 \\
13 & $1920\times{1080}$ & 2700 & 94 & 27964 & 10.3570 & 14 & $1920\times{1080}$ & 2801 &141 & 42252 & 15.0846 \\
15 & $1280\times{720}$ & 5000 & 166 & 74398 & 14.8796 & 16 & $1280\times{720}$ & 5001 & 164 & 73601 & 14.7173 \\
17 & $1280\times{720}$ & 7000 & 193 & 101165& 14.4521 & 18 & $1280\times{720}$ & 7001& 212& 103357 & 14.7632 \\
19 & $1280\times{720}$ & 7500 & 81 & 48701 & 6.4935& 20 & $1280\times{720}$ & 7501 & 86 &50832&6.7767 \\
21 & $1280\times{720}$ & 7501 & 99 & 48517 & 6.4681& 22 & $1280\times{720}$ & 7500 & 119& 65430& 8.7240 \\
23 & $1280\times{720}$ & 7500 & 115& 161220& 21.4960& 24 & $1280\times{720}$ & 7501 & 101& 150504 & 20.0645 \\
25 & $1280\times{720}$ & 5000 & 177& 93921& 18.7842& 26 & $1280\times{720}$ & 5001 & 173& 92729&18.5421\\
27 & $1280\times{720}$ & 7500 & 177& 117658& 15.6877 & 28 & $1280\times{720}$ & 7501 & 207& 140085& 18.6755 \\
29 & $1280\times{720}$ & 5001 & 168& 103983& 20.7924 & 30 & $1280\times{720}$ & 5000 & 230&119420&23.8840\\
31 & $1280\times{720}$ & 4001 & 239& 201006& 50.2389& 32 & $1280\times{720}$ & 6000 & 333& 306315& 51.0525 \\
\hline
Total & - & 104305 & 3429 & 1779866 & 17.0636  &Total & - & 107401 & 3744 & 1978955 & 18.4258 \\
\hlinew{1.1pt}
\end{tabular}
\vspace{-0.5cm}
\end{table*}

\vspace{-0.25cm}
\subsection{Data Collection}
\label{subsection:collection}
\vspace{-0.25cm}

The videos in the dataset are captured from $54$ surveillance
cameras distributed in public places. 
We collect $3888$ hours of video (over 5 TeraBytes of data in total and $72$ hours of video for each camera captured from 7:00 to 19:00 for 6 days) and select 16
representative challenging scenes based on the factors, \eg, density, diversity, occlusion, deformation, viewpoint, motion and etc. 
\footnote{The factor details are shown in the supplementary material.}
The scenes we selected are shown in Fig.~\ref{fig:scenes} and the ID numbers in the corners represent the scenarios in turn: street, crossroads, hospital entrance, school gate, park, pedestrian mall and public square.

\vspace{-0.35cm}
\subsection{Annotation}
\label{subsection:annotation}
\vspace{-0.25cm}

In order to evaluate the performance of the object tracking and behavior analysis algorithms, we proposed a clear protocol that was obeyed
throughout the annotation of the entire dataset to guarantee the consistency. The annotation rule 
\footnote{we only show the part of annotation rules due to space limited and the full rules are shown in the supplementary materials.}
includes the following
aspects:

\textbf{Class.} We annotated 7 classes of targets in urban scenarios, including \emph{pedestrian, riding, car, van, bus, tricycle}, and \emph{truck}. 
%
% There is no obvious standard division based on the appearance of diverse vehicles for each class. Therefore, the classification of vehicle during annotation is decided based on not only appearance but also general usage, \eg, SUV,MPV versus van.
% 
The queries used for each of the classes are listed as follows:
\textbf{1) \emph{pedestrian}}: single pedestrian, including the walking person, skating person and sitting person, but not including the person on the vehicle, \eg bicycle, motorcycle, scooter. Note that the pictures on the advertising boards were regarded as the background.
\textbf{2) \emph{riding}}: two-wheel vehicle with people on it, \eg, bicycle, motorcycle and scooter. The annotated bounding box surrounds the extent of both the vehicle and the person.
The \emph{person-riding} is considered as one moving individual rather than divided into vehicle and person. 
Although it looks almost the same as a pedestrian from the waist up, the person on a vehicle will be considered as a part of the vehicle and will not be annotated as \emph{pedestrian} which is different from the annotation in \cite{leal2015motchallenge,milan2016mot16}. 
The parked cycles without people will be ignored and regarded as background.
\textbf{3) \emph{car}}: four-wheel vehicle for the purpose of several-person transport, such as hatchback, sedan, SUV, MPV, taxi, jeep, convertible, etc.
\textbf{4) \emph{van}}: four-wheel medium-size passenger-car or van for the purpose of transport of a small number of cargo or used for engineering operations, such as ambulance, van, etc.
\textbf{5) \emph{bus}}: four-wheel vehicle for the purpose of taking a large number of persons, and bigger than a van, such as bus, mini-bus, coach, etc.
\textbf{6) \emph{tricycle}}: three-wheel vehicle for cargo or passenger transport, and the bounding box surrounding the extent of both vehicle and person or cargo.
\textbf{7) \emph{truck}}: four-wheel vehicle for cargo transport, such as pickup, garbage truck, lorry, fire engine, trailer, etc.

\textbf{Minimal size.}
For the case of the target size, too small targets will be ignored to make sure that the annotation accuracy is not forfeited in congested and complex scenes. 
The size threshold is defined as $60$ in pixels for the bounding box's longest side. 
If the bounding box's longest side of a target is less than $60$ in pixels in the whole trajectory, the target will be ignored. 
However, for the targets whose bounding box's longest side is less than $60$ pixels only at part of the trajectory, we still annotate these \emph{too-small} targets in order to remain the complete trajectory. 
%
% However, these \emph{too small} bounding boxes will be ignored during object detection. 
All \emph{too-small} boxes in a trajectory will ignored when evaluating the performance of the multiple-object tracking methods.

\begin{table*}[t]
\centering
\footnotesize
\caption{ Target number of each class in sample sequences. The number in brackets is percentage of each class in one sequence.}
\label{table:type_number}

\begin{tabular}{p{0.1\columnwidth}|>{\raggedleft\arraybackslash}p{0.25\columnwidth}|>{\raggedleft\arraybackslash}p{0.25\columnwidth}|>{\raggedleft\arraybackslash}p{0.25\columnwidth}|>{\raggedleft\arraybackslash}p{0.2\columnwidth}|>{\raggedleft\arraybackslash}p{0.2\columnwidth}|>{\raggedleft\arraybackslash}p{0.2\columnwidth}|>{\raggedleft\arraybackslash}p{0.2\columnwidth}}
%{|c|r|r|r|r|r|r|r|}
\hlinew{1.1pt}
  \centering Seq &  Pedestrian &  Riding &  Car &  Van & Bus & Tricycle & Truck \\
\hline
\centering 02 &  2.7k (2.36\%) &  4.0k (3.51\%) &  100.9k (\textbf{88.16\%}) & 4.4k(3.93\%) & 1.7k (1.50\%) & 0.0k (0.02\%) & 0.5k (0.50\%)\\
\centering 09 & 5.5k (6.79\%) & 11.6k (14.28\%) & 52.1k (\textbf{64.16\%}) & 1.9k (2.35\%) & 8.6k (10.65\%) & 11.0k (1.36\%) & 0.3k (0.42\%) \\
\centering 11 & 7.1k (8.49\%) & 34.8k (\textbf{41.31\%}) & 29.6k (35.18\%) & 0.0k (0.00\%) & 6.0k (7.17\%) & 2.2k (2.66\%) & 4.3k (5.20\%)   \\
\centering 18 & 59.7k (\textbf{57.47\%}) & 9.7k (9.35\%) & 24.4k (23.51\%) & 0.0k (0.07\%) & 0.7k (0.69\%) & 8.6k (8.32\%) & 0.5k (0.58\%) \\
\centering 24 & 140.6k (\textbf{93.43\%}) & 9.8k (6.57\%) & 0.0k (0.00\%) & 0.0k (0.00\%) & 0.0k (0.00\%) & 0.0k (0.00\%) & 0.0k (0.00\%) \\
\centering 32 & 251.1k (\textbf{81.98\%}) & 1.2k (0.39\%) & 54k (17.63\%) & 0.0k (0.00\%) & 0.0k (0.00\%) & 0.0k (0.00\%) & 0.0k (0.00\%) \\
\hline
\centering Overall& 1665.1k (\textbf{44.30\%}) &405.0k (10.78\%) & 1496.3k (39.81\%)& 51.1k (1.36\%) &81.8k (2.18\%) & 38.0k (1.01\%) &21.2k (0.57\%) \\
\hlinew{1.1pt}
\end{tabular}
\vspace{-0.5cm}
\end{table*}

\vspace{-0.35cm}
\subsection{Data Statistics}
\label{subsection:stastics}
\vspace{-0.25cm}

% In this section, we summarize and analyze the statistics of the dataset. The detailed statistics for each sequence is available in the
% supplementary materials.

Ultimately, we have annotated 32 sequences in total. 
The total number of annotated frames is over $200k$, resulting in $4.3$ million bounding boxes and about $7.1k$ unique trajectories, much more than the annotation in MOT challenge 2017 ($30.0k$ bounding boxes of $1.3k$ trajectories in $11.2k$ frames). 
The average density of the sequence, which means the average number of the annotated targets per frame, is almost $20$. 
Therefore, the scenes in our dataset are extremely congested and challenging. The sequences are divided into two subsets, \ie training split and test split and 
the statistics of the annotated sequences are listed in Tab.~\ref{table:sequences_included}.

\begin{figure}[t]
\centering
\subfigure[Height of pedestrians]{
\includegraphics[width=0.45\columnwidth]{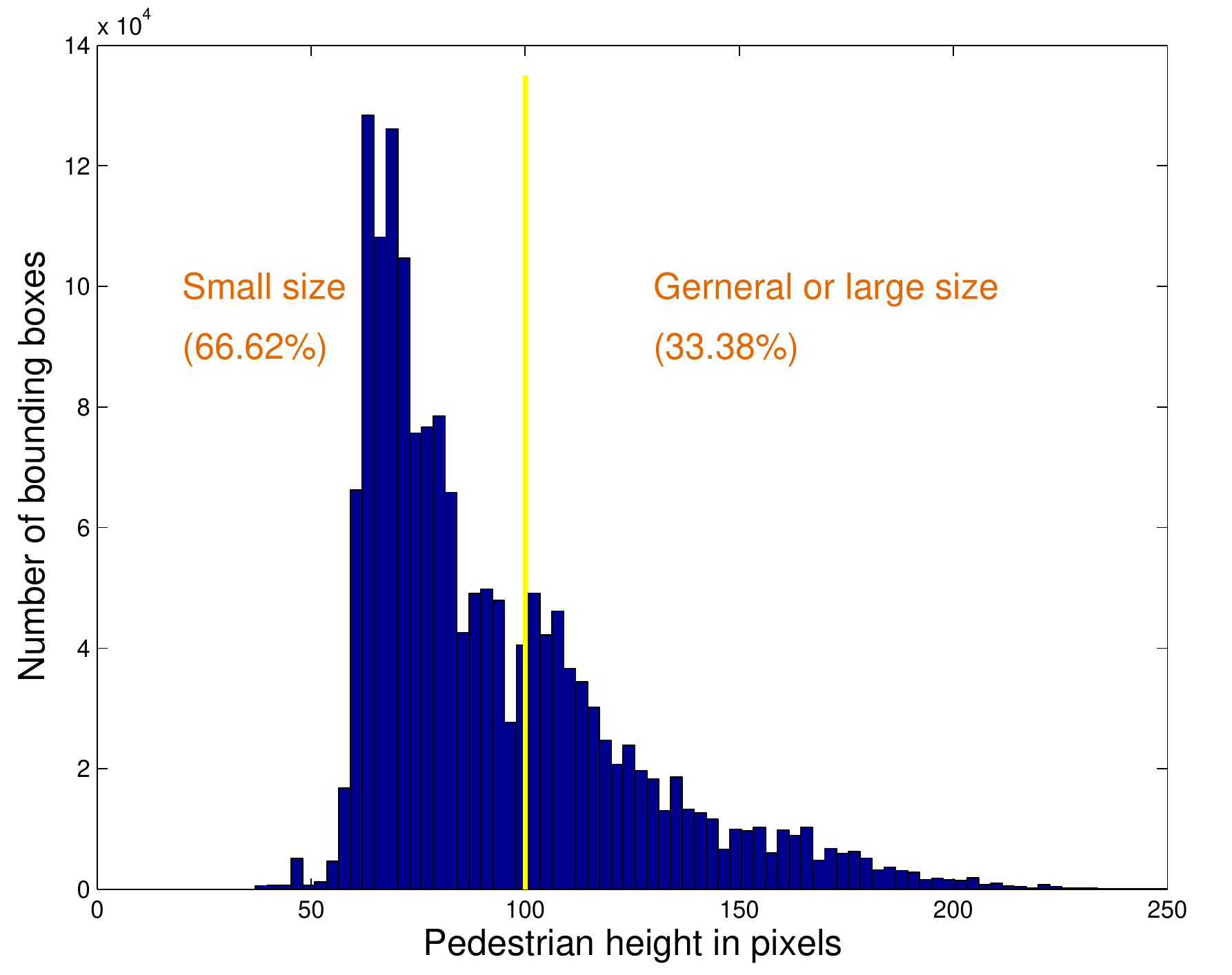}}
\subfigure[Width of cars]{
\includegraphics[width=0.45\columnwidth]{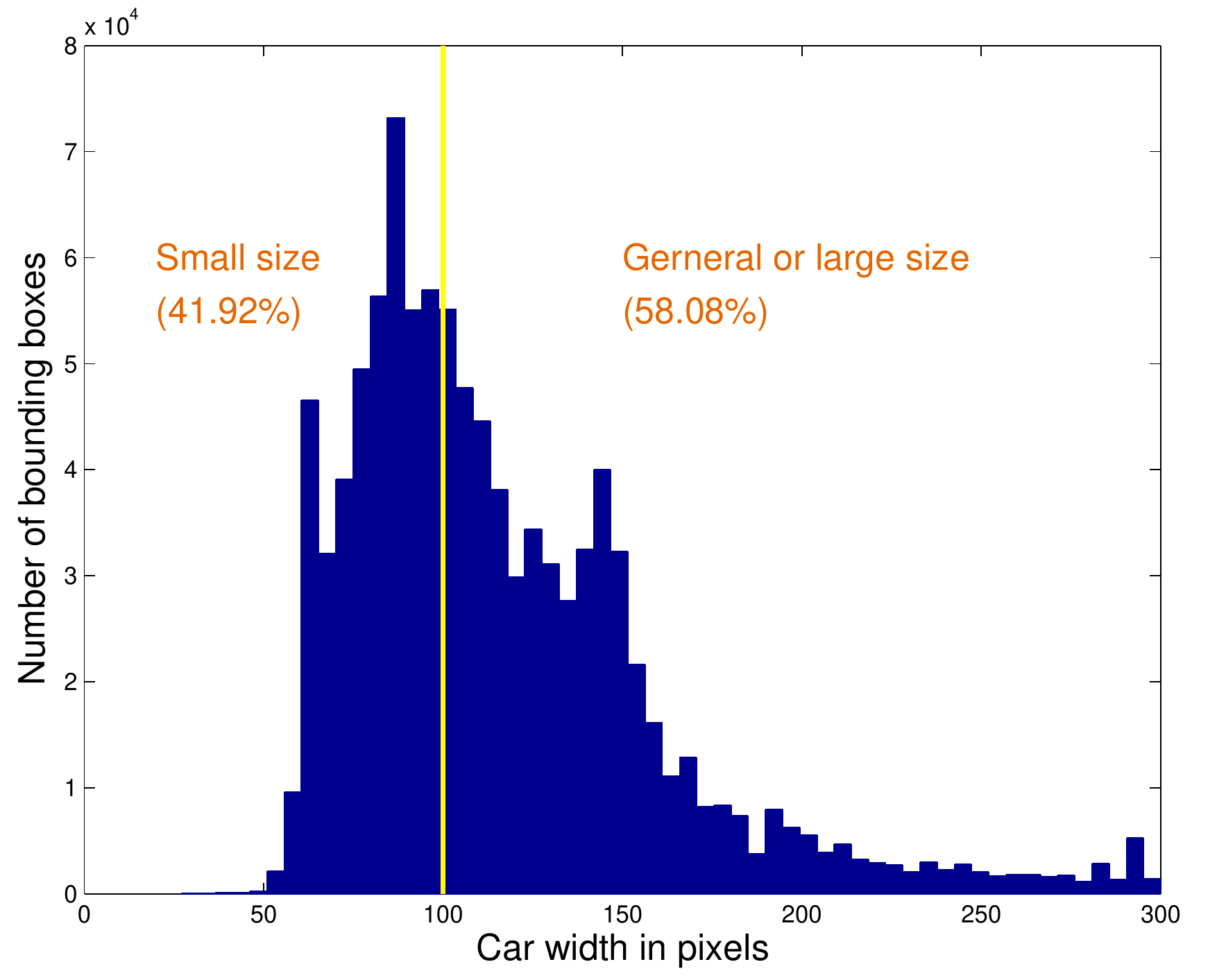}}
\vspace{-0.35cm}
\caption{Histogram of target size in pixels}
\label{fig:hist_size_length}
\vspace{-0.35cm}
\end{figure}

All the targets are grouped into 7 classes defined in Sec.~\ref{subsection:annotation} with an unique class ID for each one.
The class \emph{pedestrian} and \emph{car} are the most frequent in our dataset, with $3148.0k$ bounding boxes of \emph{car} and  \emph{pedestrian} versus $586.5k$ boxes for other classes in total. 
Furthermore, the diversity among different scenarios is relatively large. For example, the general roads include mostly cars (\eg, $83.3k$ bounding boxes versus $7.8k$ for others in total in Seq.01) while there are nearly only pedestrians in the square (\eg, $91.4k$ bounding boxes versus $2.4k$ for others in total in Seq.25).
Tab.~\ref{table:type_number} lists the class statistics of some sample sequences owing to the limited space of paper.

Most of targets in the scenes are relatively small for the reason that the surveillance cameras are a bit far from the targets for wide visual field. The small targets, \ie, the longest size length of the bounding box $100$ pixels, occupies $55.29\%$ of all targets regardless of \emph{too small} targets. 
After ignoring the \emph{too small} targets, histograms of the pedestrian height and the car width are shown in Fig.~\ref{fig:hist_size_length}.

Scale variation is one of major deformations of targets in the surveillance videos. The measure of the scale change is defined as $C=\sqrt{(H_{max}*W_{max})/(H_{min}*W_{min})}$ , where $H$ and $W$ represent the height and width of bounding box, respectively. 
The maximum and minimum size of the target is calculated based on the entire trajectory. Scale change occurs frequently in our dataset. 
For example, the size of the bounding box will change greatly when the target is approaching the camera from a distance. 
The distribution of scale change (in pixels defined above) can be grouped into three intervals by the values: \emph{small}: $1\leq{C} <1.5$, \emph{large}: $1.5\leq{C} <2.5$, and \emph{huge}: $C \geq{2.5}$. 
Each of three intervals occupies separately $31.78\%, 35.41\%$ and $32.80\%$ of all the targets. Most of the targets (trajectories) in our dataset encounter at least large scale change which makes our dataset very challenging.

Occlusion is also very frequent in the scenes which makes the object tracking very difficult. 
Nearly all the trajectories have been part-occluded and some of which are completely-occluded in our dataset. 
There are in total $596$ trajectories in which at least $38k$ bounding boxes occluded completely.

\begin{figure}[t]
\centering
% \if 0
% \subfigure[Our dataset]{
% \includegraphics[width=0.45\columnwidth]{./figures/trajectory_length.pdf}}
% \fi
\subfigure[Pedestrian]{
\includegraphics[width=0.45\columnwidth]{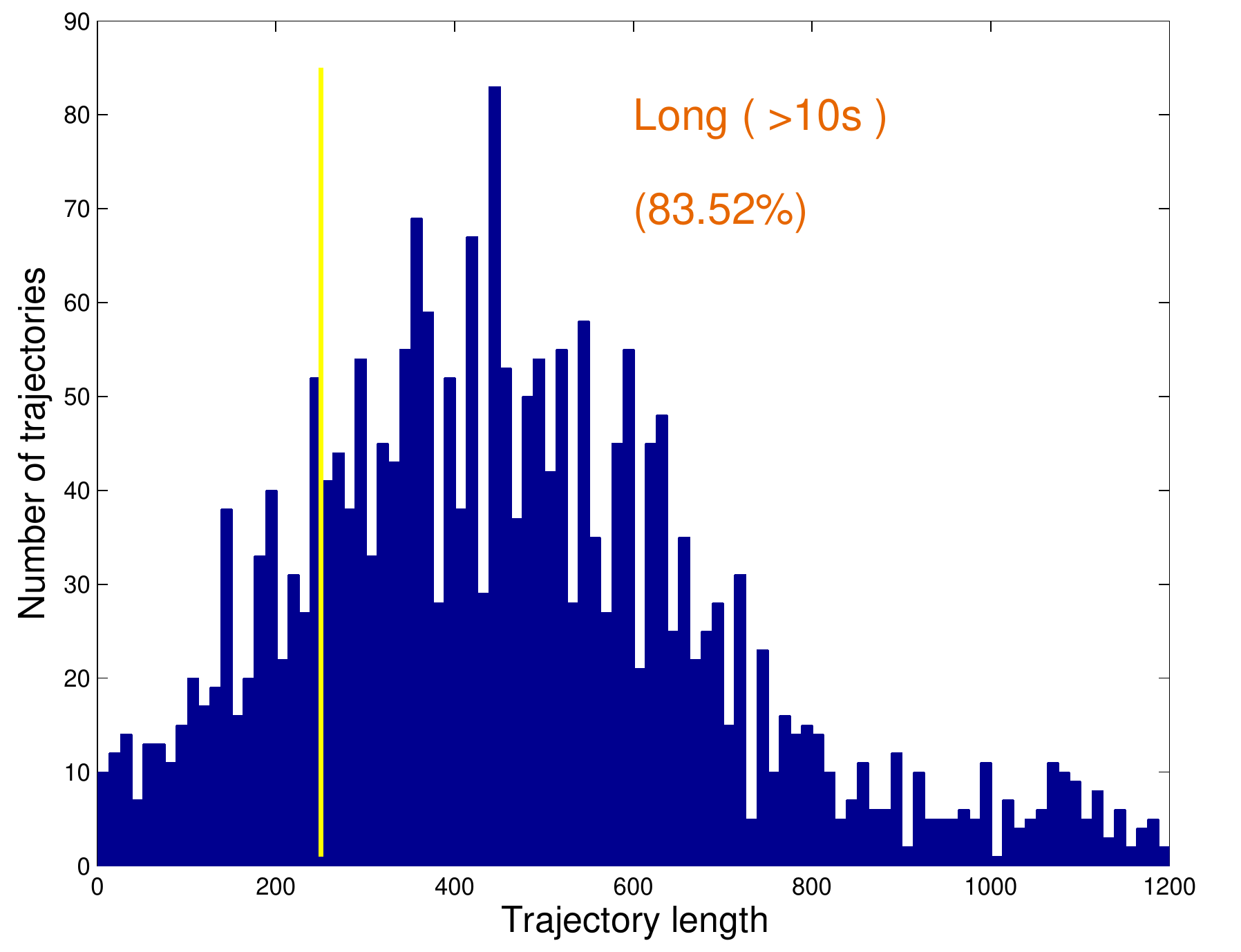}}
\subfigure[Car]{
\includegraphics[width=0.45\columnwidth]{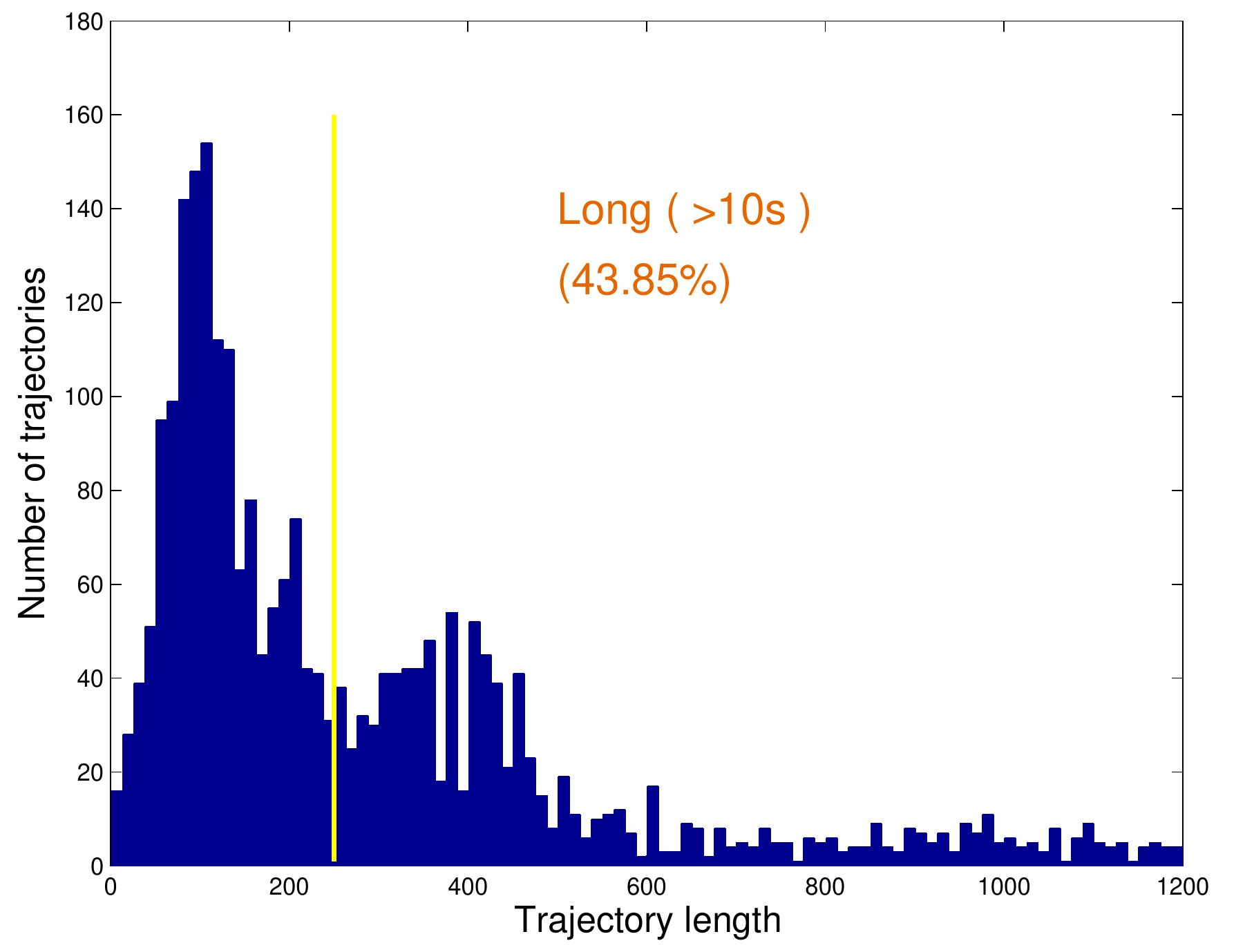}}
\vspace{-0.35cm}
\caption{Histogram of trajectory length}
\label{fig:hist_length}
\vspace{-0.35cm}
\end{figure}

The trajectory length is the time period from the appearance to disappearance of the target and depends on the speed and route of the target. The average trajectory length in our dataset is as long as $523.49$ (20.94 seconds). Generally, the trajectory length of the fast target, \eg, car, is shorter than the slower one, \eg, pedestrian. Fig.~\ref{fig:hist_length} shows the histograms of trajectory length of pedestrians and cars.

\vspace{-0.25cm}
\section{Applications and Evaluations}
\label{section:application}
\vspace{-0.35cm}

\subsection{Detection}
\label{subsection:detection}
\vspace{-0.25cm}

\begin{table*}[t]
\centering
\small
\caption{Detection results (mAP) of Faster RCNN and SSD.}
\label{table:detection_result}

\begin{tabular}{p{0.3\columnwidth}|p{0.15\columnwidth}|p{0.15\columnwidth}|p{0.15\columnwidth}|p{0.15\columnwidth}|p{0.15\columnwidth}|p{0.15\columnwidth}|p{0.15\columnwidth}|p{0.15\columnwidth}}
\hlinew{1.1pt}
Methods & Pedestrian & Riding & Car & Van & Bus & Tricycle & Truck & Average\\
\hline
SSD~\cite{liu2016ssd} & 0.6761 & 0.6768 & 0.5390 & 0.8772 & 0.6333& 0.3604 & 0.8967 & 0.6656\\
\hline
Faster RCNN~\cite{ren2015faster} & 0.6696 & 0.6671 & 0.5404 & 0.8902 & 0.6260 & 0.3572 & 0.8696 & 0.6600\\
\hlinew{1.1pt}
\end{tabular}
\vspace{-0.5cm}
\end{table*}

For each of seven classes defined in Sec.~\ref{subsection:annotation}, the goal of the object detection is to predict the bounding boxes of each target of that class in a test image (if any), with associated real-valued confidence.

All images for object detection in our dataset are sampled from the sequences by one frame per second. 
There are in total $8485$ annotated images and almost $160k$ bounding boxes for object detection, regardless of the \emph{too small} and the completely-occluded targets. 
All images are divided into train split and test split following Tab.~\ref{table:sequences_included}. Therefore, there are $4172$ annotated images for training and $4296$ images for test. 
The completely-occluded bounding boxes will be ignored in both training and testing of detection methods.
%
% The annotation of each image, which only contains the target class and the bounding box information, is provided as a corresponding XML-formatted file.
%
Faster RCNN~\cite{ren2015faster} and Single Shot MultiBox Detector (SSD)~\cite{liu2016ssd} are the most typical algorithms for object detection and achieve very good performances for object detection in PASCAL VOC~\cite{ILSVRC15}, COCO~\cite{lin2014microsoft}, and ILSVRC~\cite{deng2012ilsvrc} datasets. In our experiments, Faster RCNN and SSD will be evaluated on the performance for object detection in urban congested environments.

% {\textbf{Faster RCNN~\cite{ren2015faster}}}: In the Faster RCNN framework, the Region Proposal Network (RPN) is proposed to extract the region proposals during the detection procedure. The RPN is trained end-to-end to generate high-quality region proposals which can simultaneously predict object bounds and objectness scores at each position. It makes the detector much faster and very suitable for large scale object detection. In our experiment, Faster RCNN is based on VGG16 model~\cite{simonyan2014very}  pretrained on ILSVRC CLS-LOC dataset~\cite{russakovsky2015imagenet} and the framework is implemented as \cite{ren2015faster}. We take the default parameters as suggested by authors except that each image is resized to 720 pixels for its shorter edge. These VGG16 models are pre-trained on PASCAL VOC 2007 and the training epoch is 12.0 to avoid over-fitting.

% {\textbf{SSD~\cite{liu2016ssd}}}: The SSD model discretizes the output space of bounding boxes into a set of default boxes over different aspect ratios and scales per feature map location. The core of SSD is predicting category scores and box offsets for a fixed set of default bounding boxes using small convolutional filters applied to feature maps. The SSD model achieves very good performances. The pre-trained VGG16 models are used as the base models for the SSD framework. The input images of the SSD models are resized to $300 \times{300}$ for the compromise between speed and accuracy.

\begin{figure}[t]
\centering
\subfigure[Pedestrian Detection]{
\includegraphics[width=0.48\columnwidth]{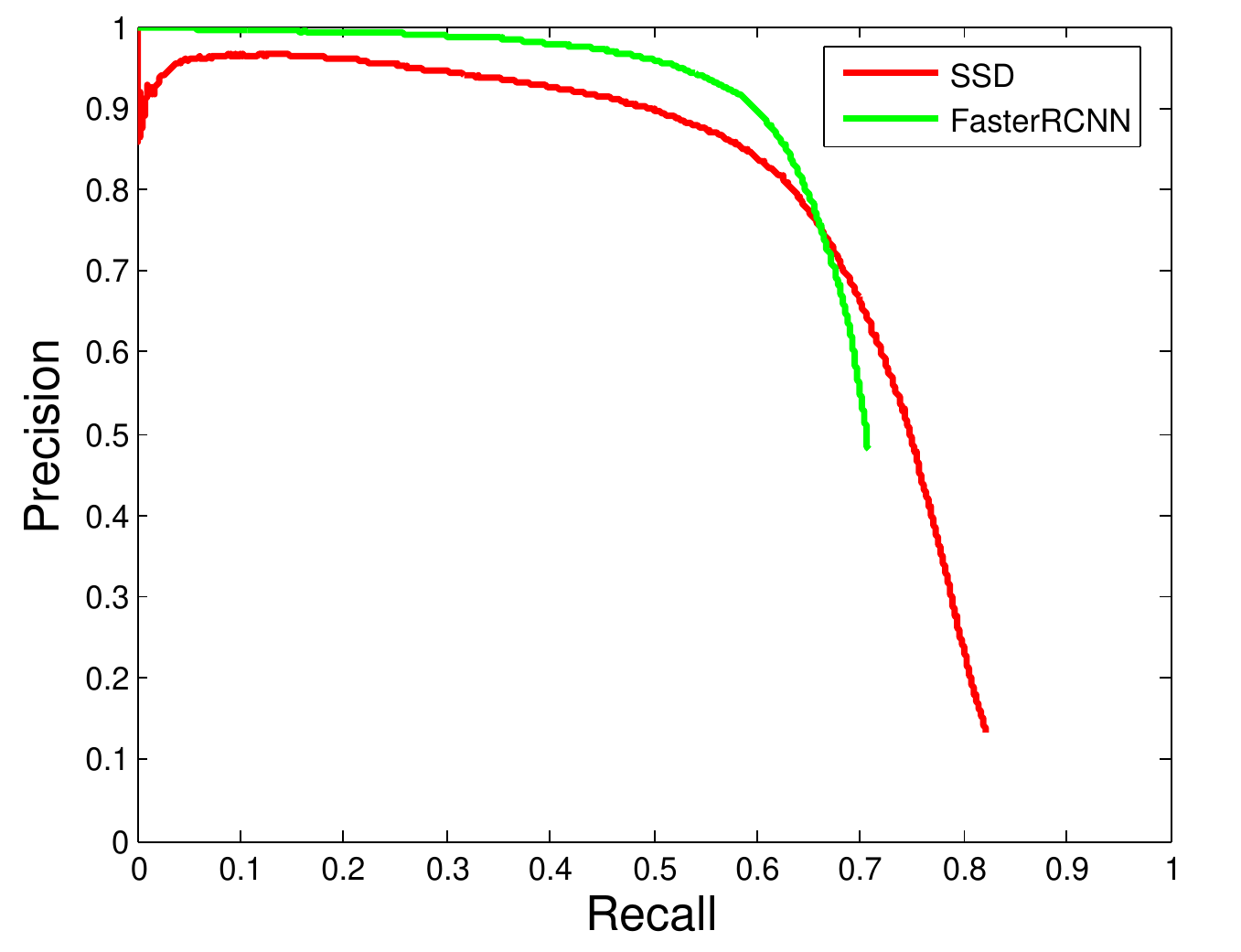}}
\subfigure[Car Detection]{
\includegraphics[width=0.48\columnwidth]{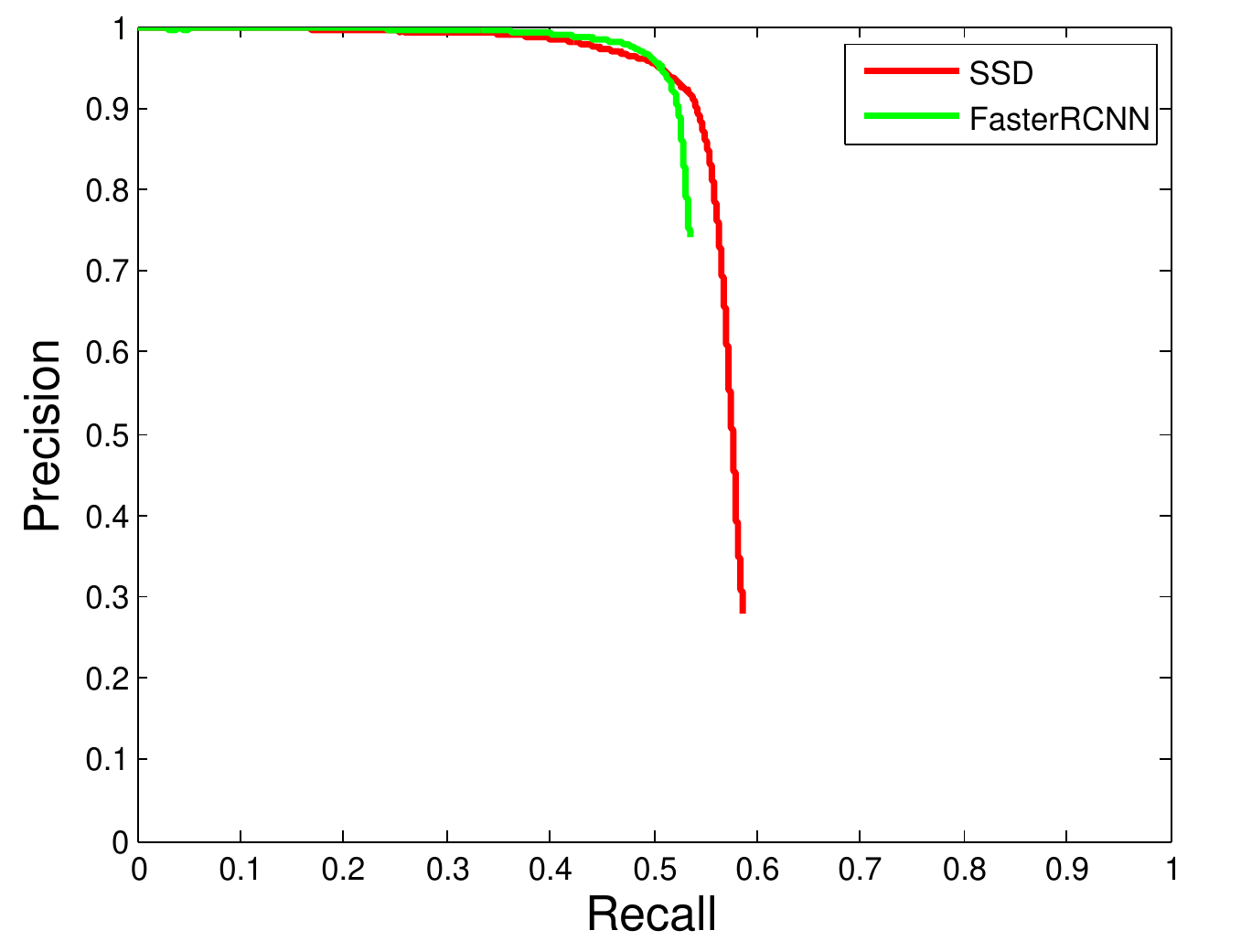}}
\vspace{-0.45cm}
\caption{Performance of detectors on pedestrians and cars. }
\label{fig:detection_result}
\vspace{-0.35cm}
\end{figure}

The evaluation of the detectors in our experiments is the same as PASCAL VOC 2007\cite{ILSVRC15} and the performances are listed in Tab.~\ref{table:detection_result}. Among all the classes of targets, we select pedestrian and car for analysis of the Precision/Recall curve as shown in Fig.~\ref{fig:detection_result}.

\vspace{-0.5cm}
\subsection{Multiple Object Tracking}
\label{subsec:mot}
\vspace{-0.25cm}

For multiple-object tracking, all the sequences are divided into training split and test split as shown in Tab.~\ref{table:sequences_included}. The train sequence and the test sequence in the same row are both captured from the same scene.
We take experiments on multiple-object tracking by using the detection results of SSD for its good performance on accuracy and speed. 
%
% For convenience and fairness of evaluation, the detection results are publicly available and the provided file of the individual sequence is formatted as comma-separated value(CSV) file in which each line represents one target instance with 10 fields as shown in Tab.~\ref{table:motformat}.
%
Evaluation metrics for multiple object tracking not only are desirable to summarize the performance into one single number to enable a direct comparison but also provide several performance estimates including information about the individual errors made by algorithms. 
Following a recent trend \cite{leal2015motchallenge} \cite{milan2016mot16}, we employ two sets of measures as the evaluation metrics for multiple-object tracking: CLEAR metrics \cite{stiefelhagen2006clear} and a set of track quality measures \cite{wu2006tracking}, \eg, MOTA and MOTP.

% \begin{table}[t]
% \centering
% \footnotesize
% \begin{tabular}{p{0.2cm}p{1.3cm}p{5.5cm}}
% %{|c|c|l|}
% \hline
% Pos & Name & Description\\
% \hline
% 1 &Frame ID& Indicate at which frame object is present\\
% 2 &Identity ID& Each trajectory is identified by a unique ID (-1 for detection)\\
% 3 &Box left(X)& Coordinate of top-left corner of bounding box\\
% 4 &Box top(Y)& Coordinate of top-left corner of bounding box\\
% 5 &Width(W)& Width in pixels of bounding box \\
% 6 &Height(H)& Height in pixels of bounding box \\
% 7 &Confidence& Indicates how confident the detector is that this instance is a pedestrian. For the ground truth and
% results, it acts as a flag whether the entry is to be considered\\
% 8 & $x$ & 3d $x$ position(-1 if not available)\\
% 9 & $y$ & 3d $y$ position(-1 if not available)\\
% 10 & $z$ & 3d $z$ position(-1 if not available)\\
% \hline
% \end{tabular}
% \caption{Data format for MOT challenge 2015}
% \label{table:motformat}
% \end{table}

\begin{table*}
\small
\caption{Quantitative results for multiple-object tracking.}
\label{table:mot_result}
\centering
\begin{tabular}{l|c|c|c|c|c|c|c|c|c|c|c}
\hlinew{1.1pt}
Methods & MOTA & MOTP & FAR & MT(\%) &ML(\%) & FP & FN & IDsw & rel. ID & FM & rel. FM  \\
% \hline
% MOTDT \\
\hline
SORT~\cite{bewley2016simple} & 37.38 & 83.01 & 0.47 & 11.04 & 28.54 & 50244 &  941580 &  18079 & 434.38 &  30650 & 736.42 \\
\hline
IOU~\cite{bochinski2017high} & 40.94 & 82.27 & 0.61 & 14.78 & 21.13 & 65344 &  847000 &  40241 & 847.46 &  40940 & 862.19 \\
\hline
TC\_ODAL~\cite{bae2014robust} &   41.05 & 78.32 &1.59& 16.25 & 29.06&  671244& 244335& 14607& 172.86 & 24662 & 291.86 \\
\hline
DP\_NMS~\cite{pirsiavash2011globally}& 20.31 & 83.95 & 3.06 & 19.68 & 38.20 & 328783 & 923420 & 5335 & 128.55 & 6184 & 149.01 \\
\hlinew{1.1pt}
\end{tabular}
\vspace{-0.5cm}
\end{table*}

In our experiments, we use several multiple-object tracking algorithms (preferring real-time methods) as baseline methods: 
1) TC\_ODAL~\cite{bae2014robust} proposed the tracklet confidence using the detectability and continuity of a tracklet, and formulated a multi-object tracking problem based on the tracklet confidence.
2) DP\_NMS \cite{pirsiavash2011globally} introduced a greedy algorithm that sequentially instantiates tracks using shortest path computations and allows one to embed pre-processing steps, such as non-max suppression, within the tracking algorithm.
3) SORT~\cite{bewley2016simple} was introduced as a pragmatic approach to multiple object tracking where the main focus is to associate objects efficiently for online applications.
4) IOU~\cite{bochinski2017high} posed a shift that enables the deployment of much simpler tracking algorithms which can compete with more sophisticated approaches at a fraction of the computational cost.
We take the default parameters as suggested by the authors and the quality measures of these baseline methods are listed in Tab.~\ref{table:mot_result}. The provided numbers may not represent the best possible performance for each method.

\begin{figure}
\begin{minipage}{0.243\columnwidth}
\centerline{\includegraphics[height=0.7\columnwidth, width=0.99\columnwidth]{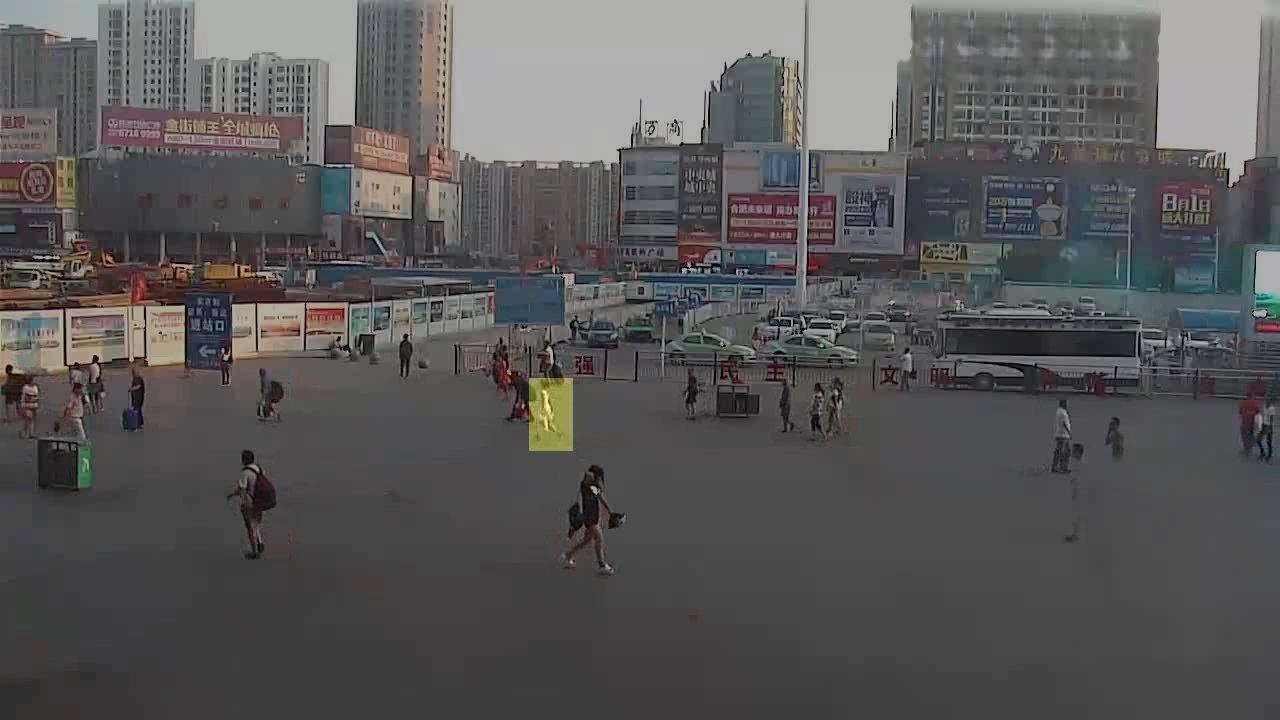}}
\end{minipage}
% \qquad
\begin{minipage}{0.243\columnwidth}
\centerline{\includegraphics[height=0.7\columnwidth, width=0.99\columnwidth]{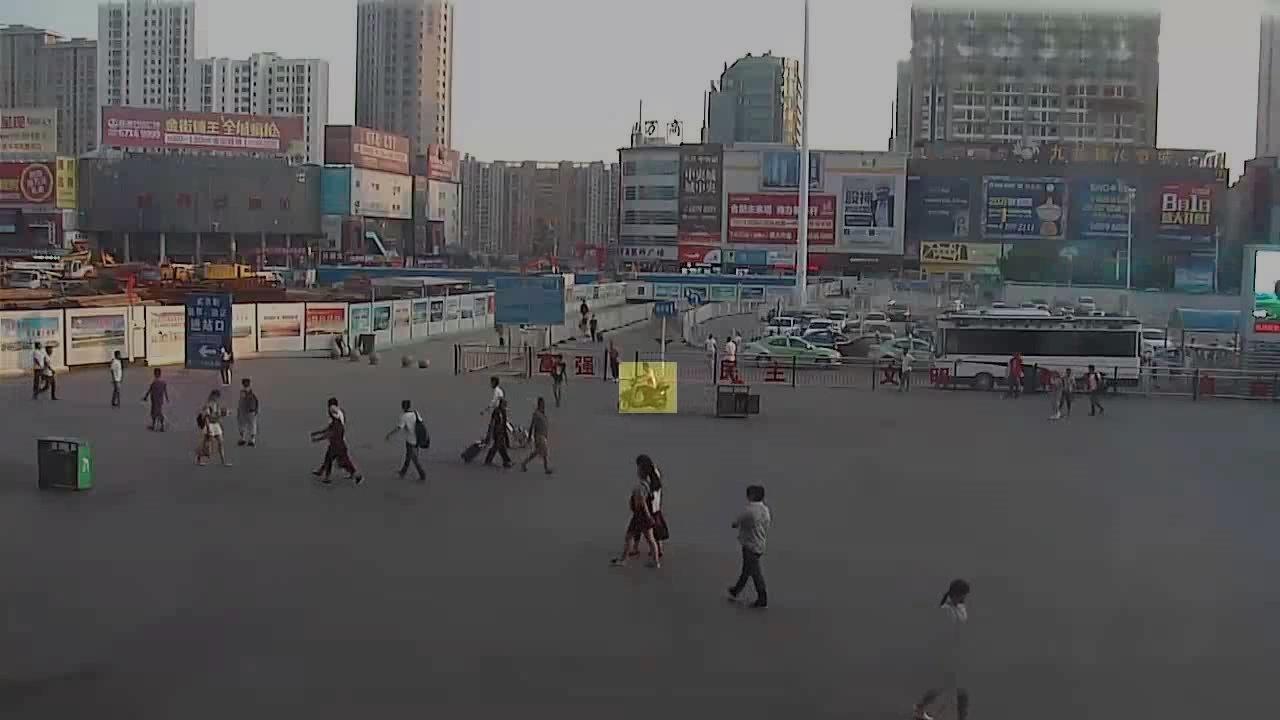}}
\end{minipage}
% \qquad
\begin{minipage}{0.243\columnwidth}
\centerline{\includegraphics[height=0.7\columnwidth, width=0.99\columnwidth]{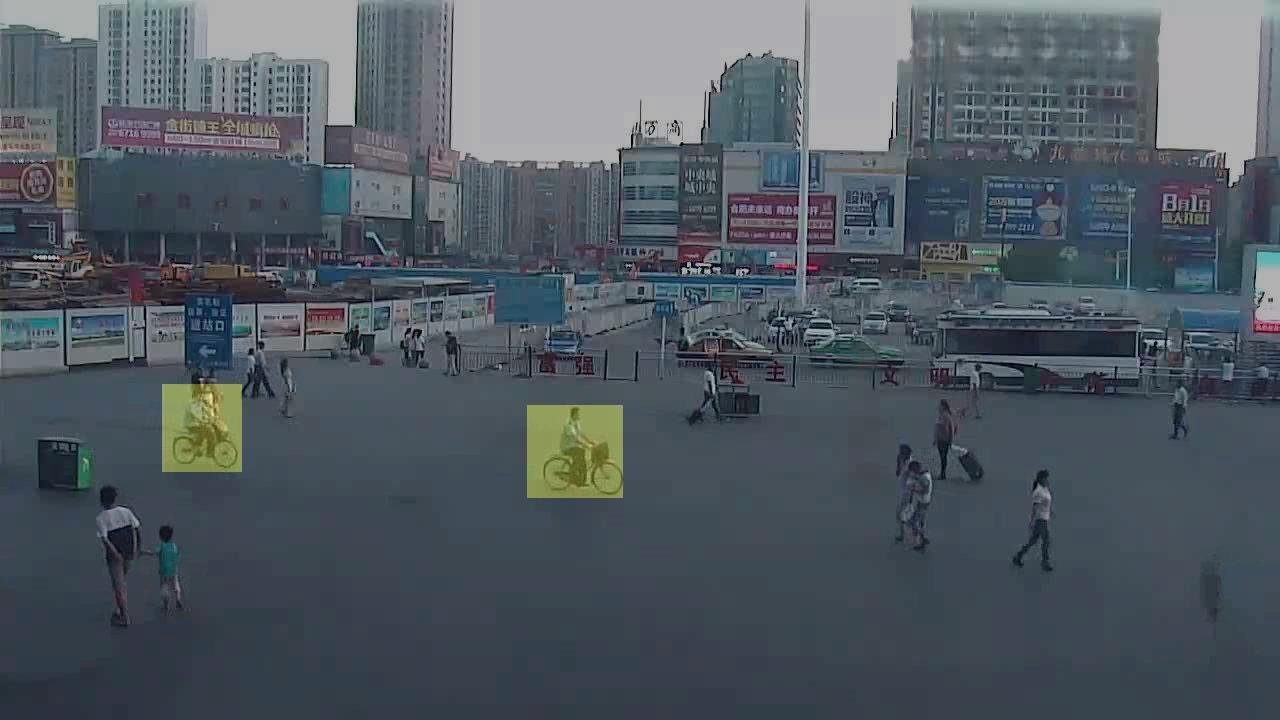}}
\end{minipage}
% \qquad
\begin{minipage}{0.243\columnwidth}
\centerline{\includegraphics[height=0.7\columnwidth, width=0.99\columnwidth]{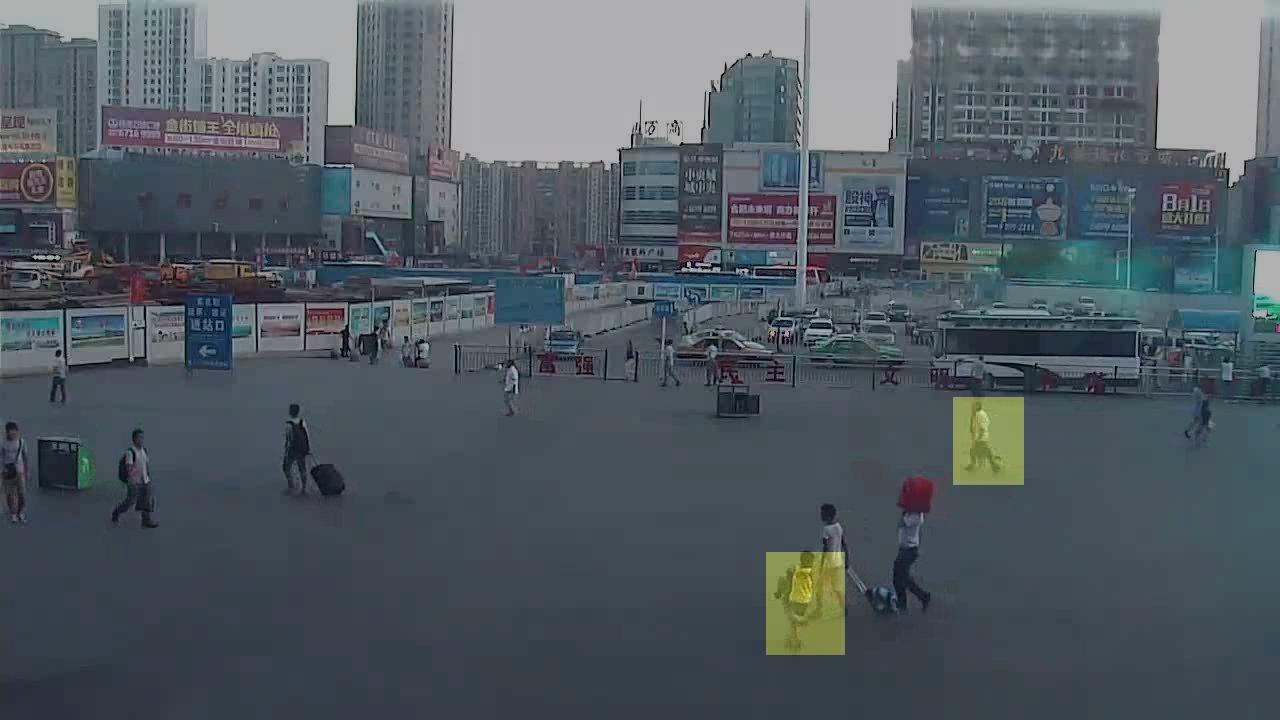}}
\end{minipage}
\\
\begin{minipage}{0.243\columnwidth}
\centerline{\includegraphics[height=0.7\columnwidth, width=0.99\columnwidth]{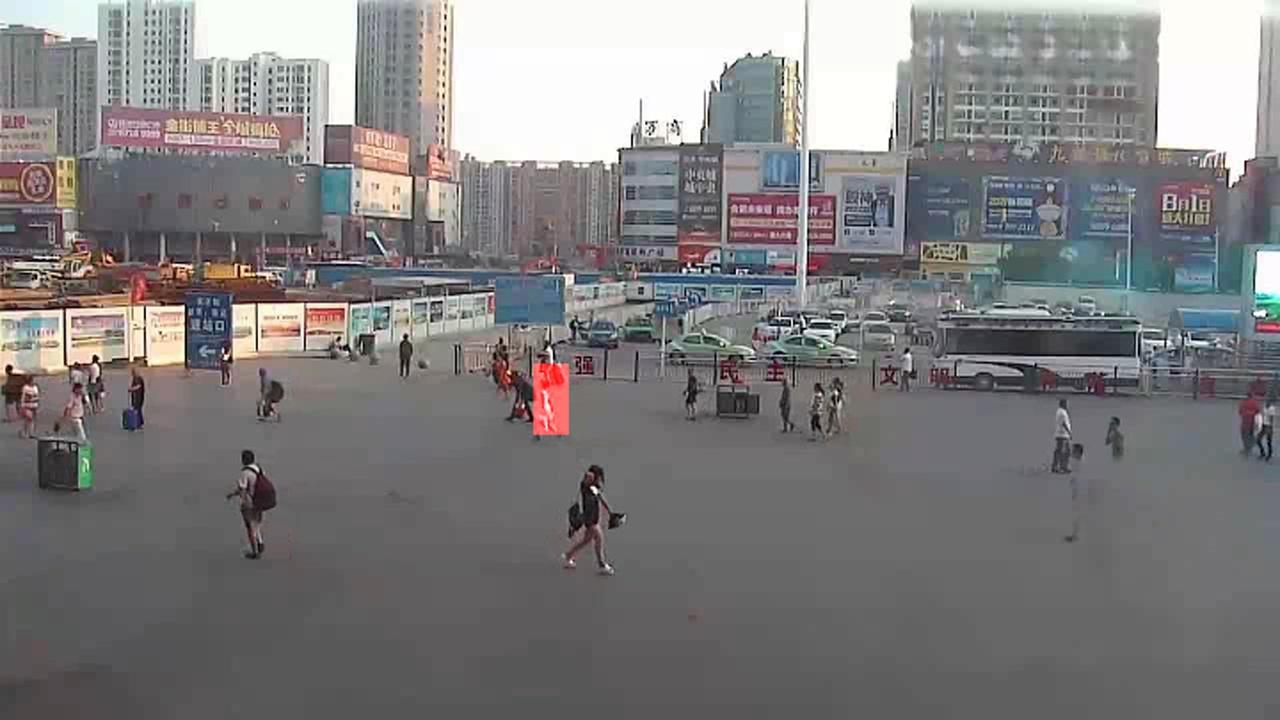}}
%\centerline{}
\end{minipage}
% \qquad
\begin{minipage}{0.243\columnwidth}
\centerline{\includegraphics[height=0.7\columnwidth, width=0.99\columnwidth]{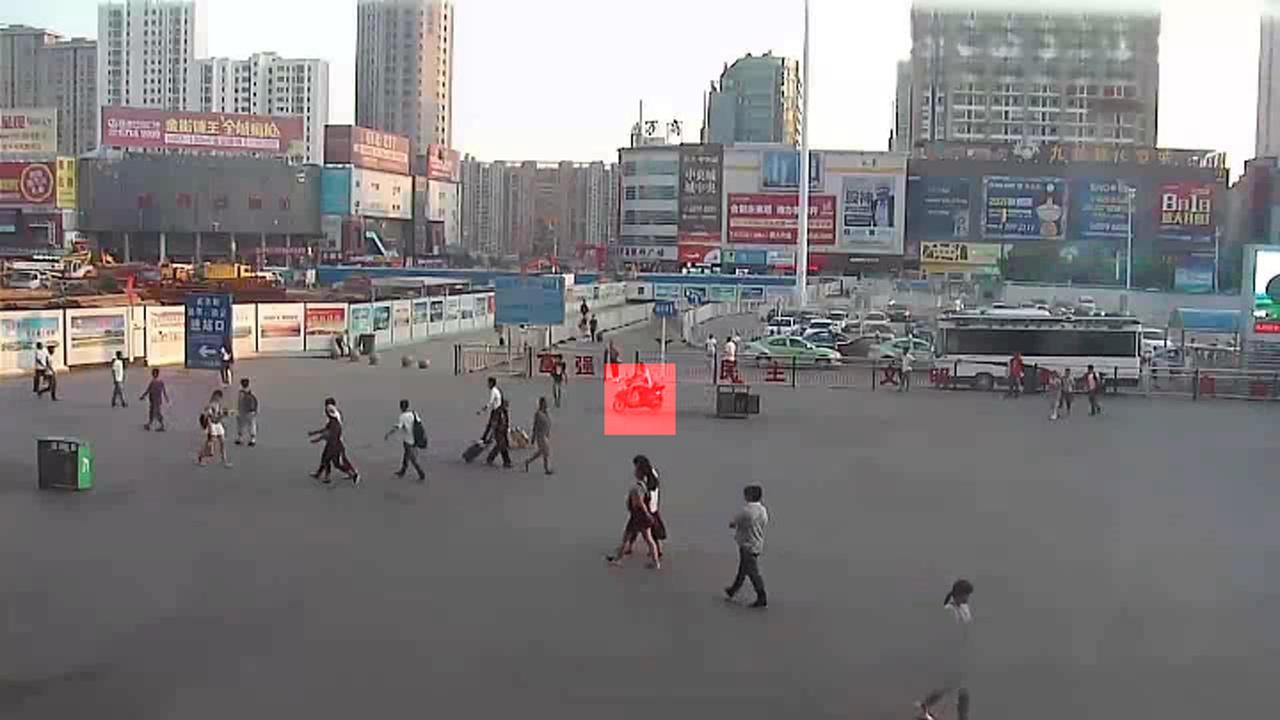}}
\end{minipage}
% \qquad
\begin{minipage}{0.243\columnwidth}
\centerline{\includegraphics[height=0.7\columnwidth, width=0.99\columnwidth]{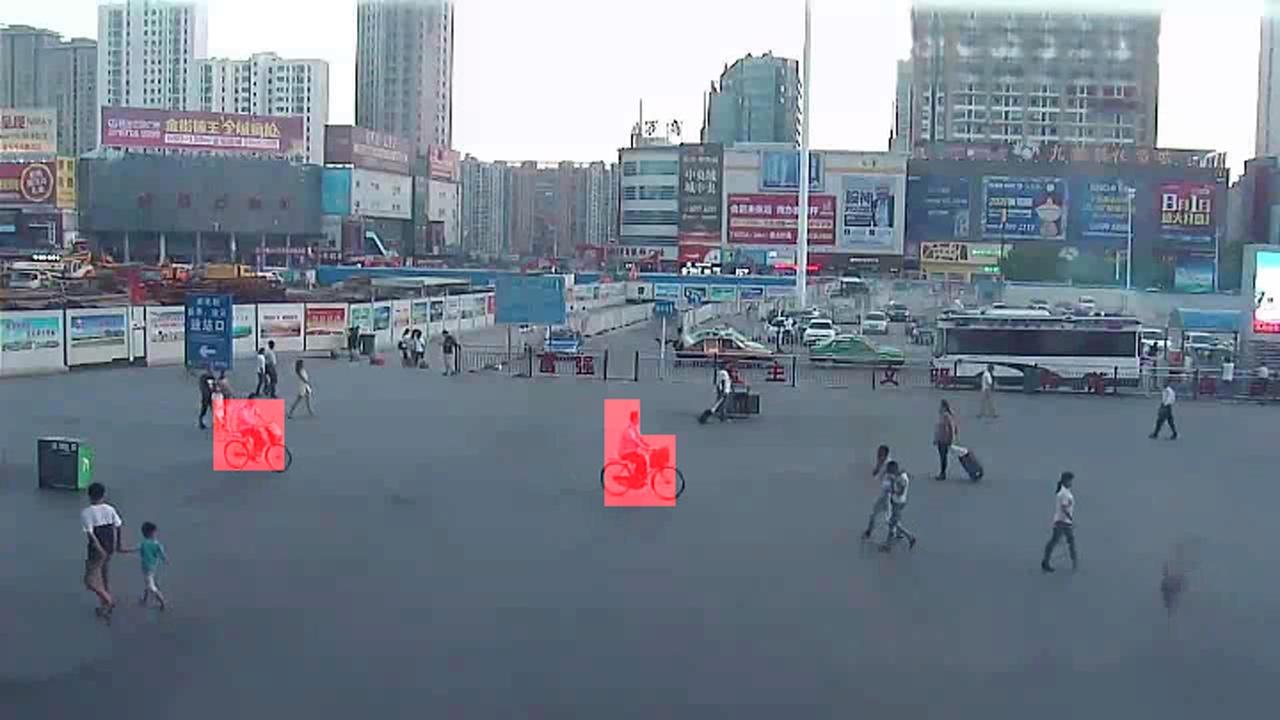}}
\end{minipage}
% \qquad
\begin{minipage}{0.243\columnwidth}
\centerline{\includegraphics[height=0.7\columnwidth, width=0.99\columnwidth]{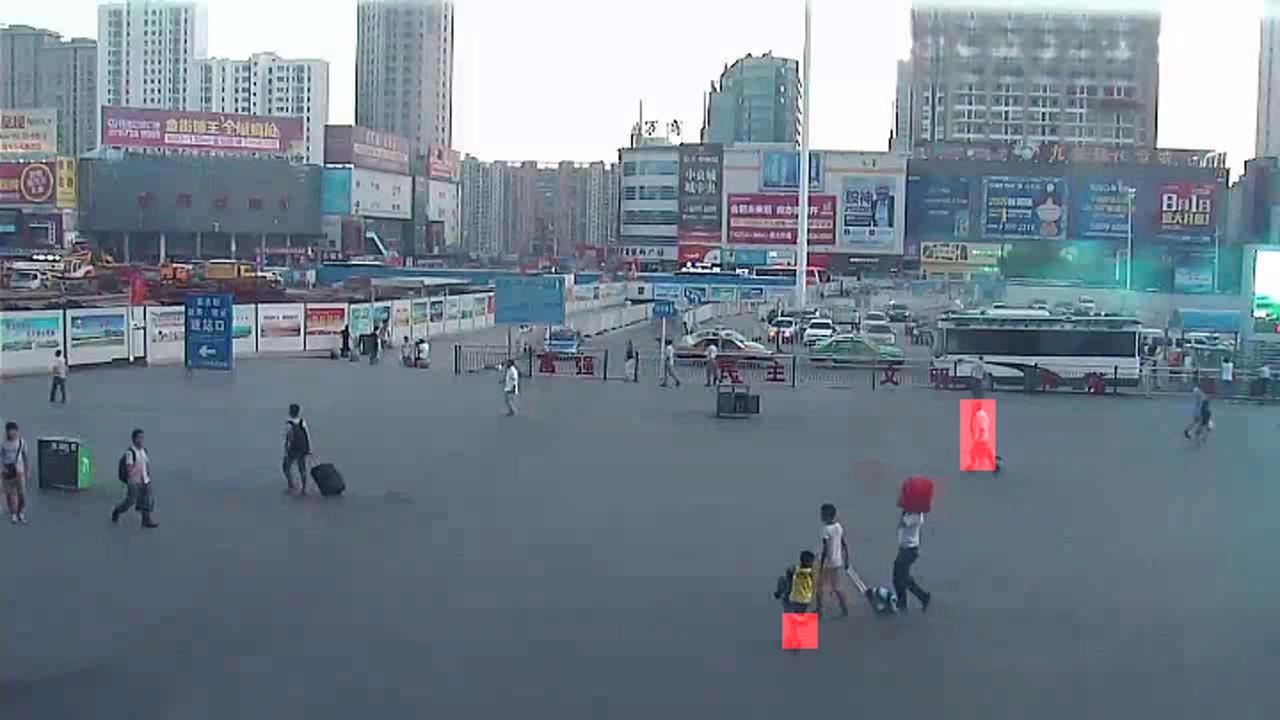}}
\end{minipage}
\vspace{-0.35cm}
\caption{Examples of different anomalies in our dataset. The top row is the ground truth and the bottom ones are the detection results of HMOF~\cite{zhuhuihui2018anomaly}.}
\label{figure:anomaly_detection}
\vspace{-0.35cm}
\end{figure}

\begin{figure}[t]
\centering
\includegraphics[width=0.8\columnwidth]{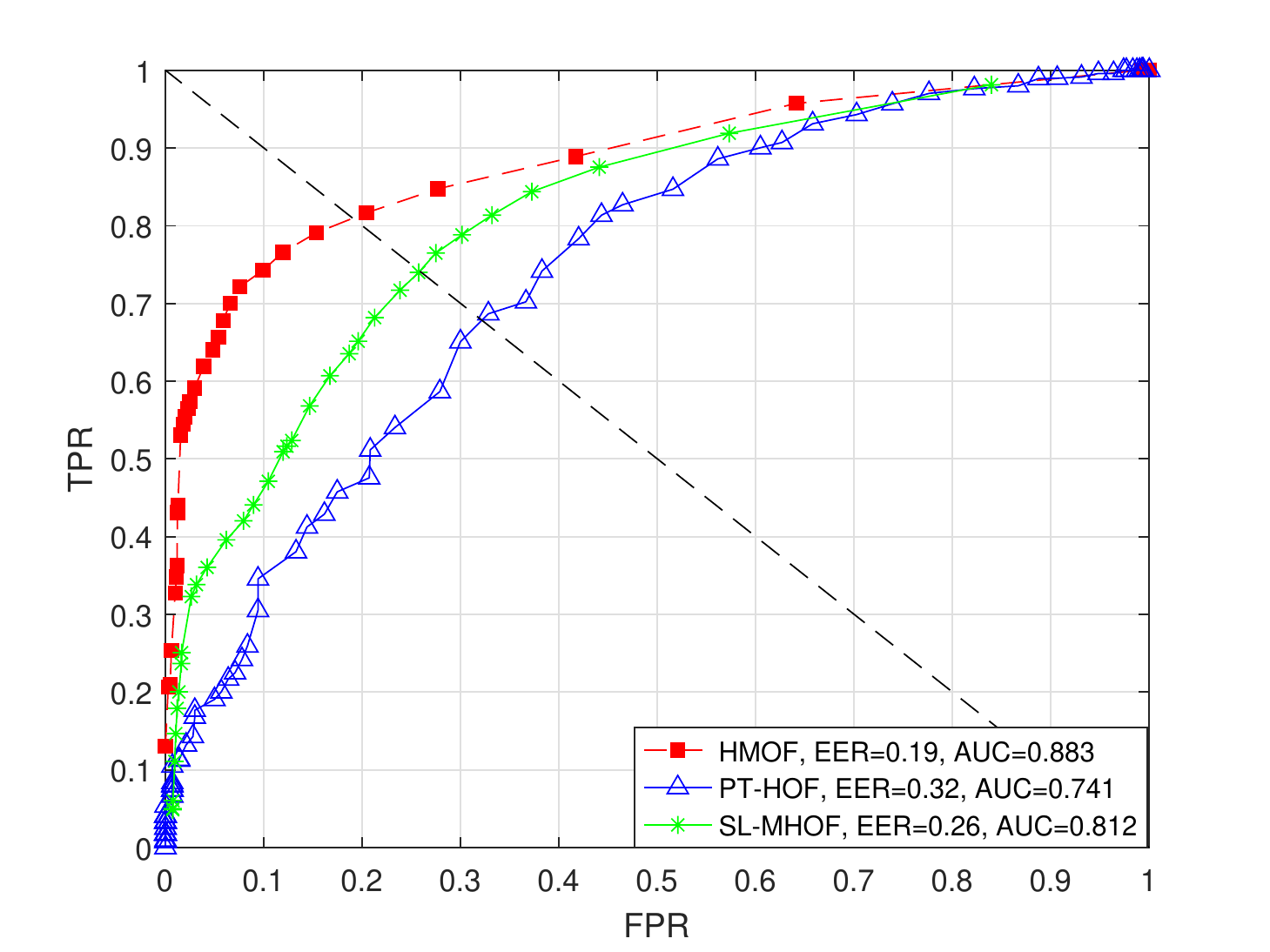}
\vspace{-0.5cm}
\caption{The ROC curves on the proposed dataset. }
\label{fig:roc_curve}
\vspace{-0.35cm}
\end{figure}

\vspace{-0.45cm}
\subsection{Anomaly Detection}
\label{subsec:behavior_analysis}
\vspace{-0.25cm}

Due to the urgent requirement of city security, anomaly detection and location is a vital task of video surveillance. 
Usually, the behaviors those rarely appeared in the videos are defined as anomaly behaviors~\cite{cong2011sparse}.
It is common practice to detect anomaly behaviors when evaluation through modeling the normal videos in the training split (zero-shot learning).
In the daily life, anomaly behaviors suffer vague definitions due to diverse scenes and relationships among objects. 
It is challenging to measure the diverse anomaly behaviors using the consistent standard.

Furthermore, we select parts of the annotated videos in our dataset for anomaly behavior detection. 
There are in total $45$ sequence videos and $200$ frames per sequence for training while $10$ sequences for test. 
The anomaly behaviors in the test split include running, jumping, bicycling, motoring and etc. As for the evaluation metrics, following~\cite{cong2011sparse,chenzhu2018anomaly,zhuhuihui2018anomaly}, we use AUC (area under curve) and EER (equal error rate) to measure the ROC curves.

In our experiments, we use several anomaly detection methods as the baselines as following:
1) Zhu \etal~\cite{zhuhuihui2018anomaly} used the HMOF features to distinguish anomaly behaviors in the videos through Gaussian Mixture Model and the visual tracking is adopted for the better detection.
2) The PT-HOF~\cite{zhao2018anomaly} were utilized to capture the fine-grained information and the consistency motion object (CMO) clusters similar point trajectories in a local region, for better anomaly localization.
3) Chen \etal~\cite{chenzhu2018anomaly} introduced a novel foreground object localization method and presented SL-MHOF, an effective descriptor modeling the local motion pattern while for appearance features, a CNN-based model is adopted.
We take the default parameters as suggested by the authors and
the ROC curves of these baseline methods are shown in Fig.~\ref{fig:roc_curve}.
The videos in our dataset are captured from the real surveillance of the congested crowds, much more complicated and challenging than other anomaly detection datasets.
Therefore, there are a long distance of the methods for real practice although they achieve excellent performances in the other datasets.

\vspace{-0.45cm}
\section{Conclusion}
\label{section:conclusion}
\vspace{-0.25cm}

In this work, we have proposed a challenging large scale urban surveillance video dataset, one of the largest and most realistic datasets, for object tracking and behavior analysis.  
The dataset consists of 16 scenes captured in 7 typical urban outdoor scenarios: street, crossroads, hospital entrance, school gate, park, pedestrian mall, and public square. 
We annotated over $200k$ video frames carefully, resulting in more than $3.7$ million object bounding boxes and about $7.1k$ trajectories. 
The proposed dataset is pretty challenging and very suitable for evaluation on object tracking and anomaly detection in urban environments.

% In the annotation procedure, we annotated a novel target class, \emph{group} defined as one unit including at least two pedestrians walking together (close location with similar velocity and direction). Furthermore, we marked the special participants of which appearance or behaviour is significantly different from the others in the traffic scenes. In the future, the dataset will be extended  for anomaly detection and crowd analysis with the annotated information. On the other hand, there will be a big data expansion of the dataset owing to annotated data occupies a small part of collected data by far.

In the annotation procedure, we annotated a novel target class, \emph{group} defined as one unit including at least two pedestrians walking together (close location with similar velocity and direction). 
In the future, the dataset will be extended  for crowd analysis with the annotated information. 
On the other hand, there will be a big data expansion of the dataset owing to the annotated data occupies a small part of collected data by far.

\vspace{0.3cm}
\noindent\textbf {Acknowledgment.}
% \subsubsection{Acknowledgment}
\vspace{0.15cm}

This work is supported by the National Natural Science Foundation of China (Grant No. 61371192), the Key Laboratory Foundation of the Chinese Academy of Sciences (CXJJ-17S044) and the Fundamental Research Funds for the Central Universities (WK2100330002, WK3480000005). \textbf{This dataset are not available due to the data license.}

\vspace{-0.25cm}
\bibliographystyle{IEEEbib}
\footnotesize
\bibliography{icme2019usvd.bbl}
\vspace{-0.25cm}
\end{document}